\providecommand{\main}{.}
\documentclass{article}

\usepackage[preprint,nonatbib]{neurips_2020}

\usepackage[utf8]{inputenc}
\usepackage{hyperref}
\usepackage[
            style=alphabetic,
		    sorting=debug,
		    maxbibnames=9,
		    maxcitenames=1,
            giveninits=true,
		    language=english]{biblatex}

\usepackage{mathtools}
\usepackage{mdframed}
\usepackage{mathrsfs}
\usepackage{amsmath}
\usepackage{amssymb}
\usepackage{amsthm}
\usepackage{cleveref}
\usepackage{tikz}
\usetikzlibrary{matrix,arrows}
\usepackage{tikz-cd}
\usepackage{csvsimple,longtable,makecell,booktabs}

\newtheorem{theorem}{Theorem}[section]
\theoremstyle{plain}
\newtheorem{remark}[theorem]{Remark}

\usepackage{dsfont,bbold}

\usepackage{algorithm,algpseudocode}
\usepackage{colonequals}
\usepackage{enumitem}
\usepackage{graphicx}
\usepackage{comment}
\usepackage{contour}
\usepackage[linecolor=white,backgroundcolor=white,bordercolor=white,textsize=tiny]{todonotes}
\let\todon\todo
\renewcommand{\todo}[1]{\todon[inline,color=green!40]{\color{blue}{#1}}}
\usepackage[skip=2pt,font=scriptsize]{caption}

\usepackage{\main/style}

\graphicspath{{\main/images/}{images/}}

\title{FRUITS: Feature Extraction Using Iterated Sums\\for Time Series Classification}
\author{
    Joscha~Diehl\\
    Universität Greifswald\\Institut für Mathematik und Informatik\\
    \texttt{joscha.diehl@uni-greifswald.de} \\
    \And
    Richard~Krieg \\
    Universität Greifswald\\Institut für Mathematik und Informatik\\
    \texttt{s-rikrie@uni-greifswald.de}\\
}

\nocite{*}

\newcommand\DEF[1]{\textbf{#1}}
\newcommand\magenta[1]{{\color{magenta}#1}}

\def\w#1{{\color{cyan}\mathtt{#1}}}

\renewcommand{\itsum}[2][{}]{\ISS_{#2}^{#1}}
\renewcommand{\itsumu}[2][{}]{\underline{\ISS}_{#2}^{#1}}

\renewcommand\cos{\mathsf{cos}}
\renewcommand\sin{\mathsf{sin}}

\begin{document}

\maketitle

\begin{abstract}
We introduce a pipeline for time series classification that extracts features based on the iterated-sums signature (ISS) and then applies a linear classifier.  These features are intrinsically nonlinear, capture chronological information, and, under certain settings, are invariant to time-warping.  We are competitive with state-of-the-art methods on the UCR archive, both in terms of accuracy and speed.

We make our code available at \url{https://github.com/irkri/fruits}.
\end{abstract}

\tableofcontents 

\section{Introduction}

Time series classification is a fundamental problem in data science.
It has a wide range of applications \cite{kampouraki2008heartbeat,mubashir2013survey,zheng2014time}.

Deep learning methods have made significant progress in this area
\cite{wang2017time,ismail2019deep}.
But more ``classical'' methods are still competitive / SOTA,
\cite{bagnall2017great,DPW2020,Mid2021,tan2022multirocket,dempster2023hydra,MSB2023}
with the added benefits of simplicity and speed.

In this paper, we present a pipeline for time series classification
that extracts features based on the iterated-sums signature (ISS)
\cite{DEFT2020,DEFT2022} and then applies a linear classifier.
The pipeline is called FRUITS (Feature Extraction Using Iterated Sums),
and its two-stage approach is similar to the one of the Rocket
\cite{DPW2020} pipeline.

The substantial difference is that \textbf{FRUITS' iterated sums
are nonlinear to begin with}, whereas Rocket uses random \emph{linear} convolutions
(and achieves nonlinearity through pooling operations before the linear classifier).

We achieve competitive results on the UCR archive \cite{Dau2019},
and, for choices of hyperparameters, time-warping invariance.

\paragraph{Iterated sums and integrals in data science}

Let $x:\N\to\R$ be a one-dimensional time series.
Iterated sums are (mostly nonlinear) transformations of $x$
obtained by summing certain polynomial expressions of its past values.
Simple examples are
\begin{align*}
    \sum_i x_i, \quad \sum_{t_1<t_2} x_{t_1} x_{t_2}, \quad \sum_{t_1<t_2} x_{t_1}^2 x_{t_2}.
\end{align*}
Note that the first sum here is linear in $x$ (it is the only linear ``iterated'' sum).
It is well-known that a simple sum of this type is invariant to arbitrary permutations of the time steps
(leading, for example, to their use in deep sets \cite{zaheer2017deep}).
We note that the second sum, which is an honest \emph{iterated} sum,
namely
\begin{align*}
    \sum_{t_2} \left( \sum_{t_1 < t_2} x_{t_1} \right) x_{t_2},
\end{align*}
is also invariant to arbitrary permutations of the time steps.
The third one is \emph{not} and it therefore truly captures some sort of
``chronological'' information.

Iterated sums in general \emph{are} invariant to arbitrarily inserting zeros
into a sequence.
This invariance played a key role in their development and is one of their
most important properties, \cite{DEFT2020,DEFT2022}.
When applying iterated sums to the \emph{increments} $\delta x_i = x_i - x_{i-1}$,
it yields invariance to ``stuttering'' or repetition of values,
and hence an invariance to time-warping.
Such time-warping invariance is important in many applications,
%
as evidenced by the extensive literature on dynamic time warping (DTW)
\cite{berndt1994using,yi1998efficient,keogh2005exact,cuturi2017soft}.

Another perspective on iterated sums is considering them a discretization of \emph{iterated integrals}.
These features of continuous-time curves date back to Chen's work on the homology
of path spaces \cite{chen1957integration} and have been used in the last decades
in control theory \cite{fliess1981fonctionnelles},
rough path analysis \cite{lyons1998differential}
and, more recently, data science \cite{xie2017learning,chevyrev2018persistence,diehl2019invariants,kidger2019deep,kiraly2019kernels,kidger2020neural,cuchiero2021expressive}.
We note that applying iterated integrals to a one-dimensional
time series results in trivial features (owing to the fundamental theorem of calculus).
This well-known problem is usually circumvented
by first lifting the time series to a higher-dimensional space,
for example by adding a time-dimension.
The ISS does \emph{not} necessitate such a lift, as it provides
a huge variety of nonlinear features already in one dimension.

\paragraph{Contributions}

We provide, for the first time, a comprehensive study of the use of the
iterated-sums signature (ISS) as a feature-extraction method for time series classification.
We restrict to a ``classical'', non-deep learning setting:
the features are not learned, but computed
from the data and then used in a linear classifier.
This is accompanied by a fast implementation of the ISS,
in a well-documented and unit-tested Python package, \url{https://github.com/irkri/fruits}.

We show that our method is competitive with state-of-the-art methods on the UCR archive,
while providing additional benefits such as being deterministic,
and providing for certain choices of hyperparameters, time-warping invariance.

Moreover, we develop the ISS in the following directions:
\begin{itemize}

    \item
        We introduce a weighting scheme for the ISS that penalizes/boosts
        summands depending on the distance of their indices.
        This idea appeared in \cite{Kri2021} for the first time,
        and we extend it to new weighting schemes
        that bear close resemblance
        to positional encodings in NLP \cite{vaswani2017attention}.

    \item
        In the arctic semiring, in addition to the
        iterated sums, which yield certain \emph{values} at global maxima/minima,
        we provide a linear-in-time algorithm to also obtain the \emph{indices}.

\end{itemize}

\paragraph{Outline}

In \Cref{sec:iss}, we introduce the ISS and its generalization to semirings.
In \Cref{sec:pipeline} we describe our pipeline, that uses the ISS as the central
feature extractor.
In \Cref{sec:experiments}, we present our experiments.

The appendix contains a more detailed introduction to the ISS
and an algorithm to obtain the indices of the global maxima/minima
that is mentioned in \Cref{ss:semirings}.

\section{The iterated-sums signature}
\label{sec:iss}

$\N = \{0,1, \dots \}$ denotes the non-negative integers,
and $\N_{\ge 1} = \{1, 2, \dots \}$ the positive integers.

We will denote a $d$-dimensional time series by a lowercase letter $x\in\tset{d}$ and interpret it as a function
on the natural numbers,
\begin{align*}
    \tset{d} :=\{x\mid x:\N_{\ge 1} \to\R^d\}.
\end{align*}
The collection of time series of length $T$ will be denoted by
\begin{align*}
    \tset[T]{d} :=\{x\mid x:\{1,\dotsc, T\}\to\R^d\}.
\end{align*}
We write $x_t=x(t)\in\R^d$ for a single time step $t\in\N_{\ge 1}$. Entries of this $d$-dimensional vector
are accessed using a superscript $x_t^{[j]}$ for $j\in\{\w1,\dotsc,\w d\}$.
We extend this notation to formal, commutative monomials of the indices, e.g.
$x_t^{[\w 1^2 \w3]}=x_t^{[\w 1]}x_t^{[\w 1]} x_t^{[\w 3]}$.

\subsection{Iterated sums over the reals}\label{sec:iss_def}

The \DEF{iterated-sums signature (ISS)} was first introduced, in an algebraic framework,
in \cite{DEFT2020}.
We sketch its construction here, more details can be found in  Appendix~\ref{app:iss}.

The ISS consists of polynomial expressions in the time series' values,
indexed by \DEF{words} $w=[a_1]\dotsc[a_p]$.
Here, a \DEF{letter} $[a_i]$ is a non-constant monomial in the dummy
variables $\w{1}, ..., \w{d}$
(in other words $[a_i] \in \N^d$, with at least one non-zero entry).
The total number of those variables in a word $w$ is called its \DEF{weight} $\abs{w}$.
For a fixed word $w$ we obtain a new (one-dimensional) time series,
$\itsum{w}(x) \in \tset{1}$,
where the value at each time step is an iterated sum of the input signal's past
as follows
\begin{align}\label{eq:iss_coeff}
    \itsum{w}(x)_t
    \coloneqq\sum_{0<t_1<\dotsc< t_p\leq t}
        x_{t_1}^{[a_{1}]}\cdots x_{t_p}^{[a_{p}]}.
\end{align}

For example, for a two-dimensional time series $x\in\tset[T]{2}$ and the word $w=[\w1^2\w2][\w2^3]$ we have
\begin{align*}
    \itsum{[1^22][2^3]}(x)_t=\sum_{0<t_1<t_2\leq t} x_{t_1}^{[\w1^2\w2]}\cdot x_{t_2}^{[\w2^3]}=
        \sum_{0<t_1<t_2\leq t}
            \left(x_{t_1}^{[\w1]}\right)^2x_{t_1}^{[\w2]}\left(x_{t_2}^{[\w2]}\right)^3.
\end{align*}
i.e. formal products in $a_i$ translate to products of the corresponding dimensions at
coinciding time step in the iterated sum.
The ISS therefore allows different time steps as well as
dimensions in a time series to interact, non-linearily, with each other.

\subsection{Iterated sums over semirings}
\label{ss:semirings}

Semirings are algebraic structures that generalize rings by dropping the requirement of additive inverses.
They appear for example in theoretical computer science, where they are used in automata theory
and in describing certain dynamic programming algorithms, \cite{mohri2002semiring}.

\citeauthor{DEFT2022} \cite{DEFT2022} show that the ISS can actually be defined over arbitrary
commutative semirings,
another fact that distinguishes it from iterated integrals.
A \DEF{commutative semiring} $\semiring$ is a tuple $(S,\oplus,\odot,\zero,\one)$ where
\begin{itemize}
    \item $S$ is a set,
    \item $\oplus: S\times S\to S$ is associative, commutative and $\zero\oplus s=s$ for
    all $s\in S$,
    \item $\odot: S\times S\to S$ is associative, commutative and $\one\odot s=s$ for
    all $s\in S$ and
    \item $\zero\odot s=s\odot\zero=\zero$ for all $s\in S$.
\end{itemize}
Every commutative ring is a commutative semiring.
For the commutative ring of real numbers $\reals=(\R,+,\cdot,0,1)$ the iterated-sums signature is
already introduced in Section~\ref{sec:iss_def}.
Replacing the standard sum and product of that definition
by the operations in a commutative semiring $\semiring$,
we obtain the following definition
\begin{align}\label{eq:itsum}
    \begin{split}
    \itsum[\semiring]{w}:\tset{d}&\to\tset{1} \\
    \itsum[\semiring]{w}(x)_t&=
        \bigoplus_{0<t_1<\dotsc< t_p\leq t}
        x_{t_1}^{\odot [a_{1}]}\odot\cdots\odot x_{t_p}^{\odot [a_{p}]}
    \end{split}
\end{align}
for some word $w=[a_1]\dotsc [a_p]$. The symbol $\odot$ in the exponent $x_{t_j}^{\odot [a_{j}]}$
now highlights that a formal product of dimension indices in $a_j$ translates to repeatedly
applying $\odot$.

\paragraph{Arctic iterated sums}
The canonical example of a commutative semiring that is \emph{not} a ring
is the \DEF{arctic} (or \DEF{max-plus}) semiring
${\arctic=(\R\cup\{-\inf\},\max,+,-\inf,0)}$.\footnote{Equivalently, one can use
the \DEF{tropical} semiring ${\tropical=(\R\cup\{\inf\},\min,+,\inf,0)}$.}
As an example, let $w=[1][1^2][23]$ and $x\in\tset[T]{3}$. We then have,
\begin{align*}
    \itsum[\arctic]{w}(x)_T
        &=\max_{0< t_1< t_2< t_3\leq T}
            x_{t_1}^{\odot [1]}+x_{t_2}^{\odot [1^2]}+x_{t_3}^{\odot [23]}\\
        &=\max_{0< t_1< t_2< t_3\leq T}
            x_{t_1}^{[1]}+2x_{t_2}^{[1]}+x_{t_3}^{[2]}+x_{t_3}^{[3]}.
\end{align*}

A real-valued time series can be considered as taking values in the arctic semiring.
Calculating the corresponding (arctic) iterated sums leads to features
that are quite different from the (classical, real) iterated sums.
The latter are smooth, polynomial expressions
whereas the former are piecewise linear expressions of cumulative maxima.

As noted in \cite{DEFT2022}, such expressions
are not time-warping invariant (but, of course, invariant to insertion of zeros,
which in this case corresponds to the insertion of $-\infty$).
One obtains time-warping invariant features by using non-strict inequalities
for the indices. We are thus led to define
\begin{align}\label{eq:itsum_semiring}
    \begin{split}
    \itsumu[\arctic]{w}:\tset{d}&\to\tset{1} \\
    \itsum[\arctic]{w}(x)_t&=
        \scalar{w,\iss[\arctic]{0,t}(x)}=\bigoplus_{1\leq t_1\leq\dotsc\leq t_p\leq t}
            x_{t_1}^{\odot [a_{1}]}\odot\cdots\odot x_{t_p}^{\odot [a_{p}]}.
    \end{split}
\end{align}

\begin{figure}
    \includegraphics[width=\textwidth]{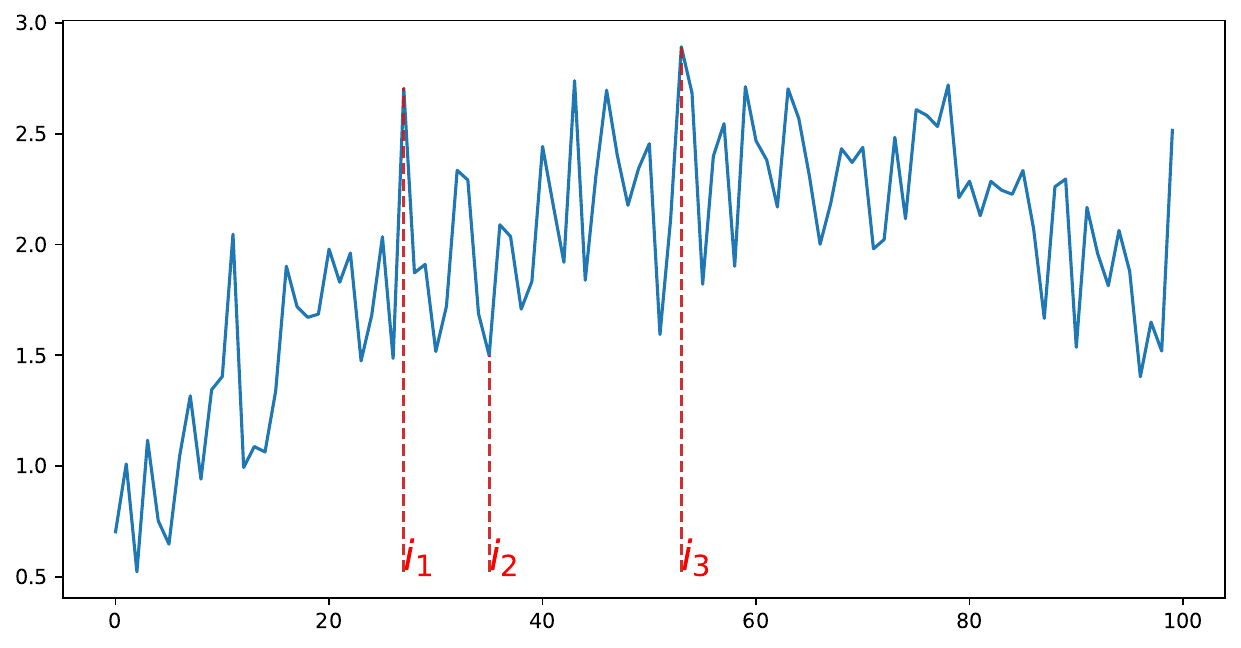}
    \caption{Indices in the arctic iterated sum $\itsum[\arctic]{[1][1^{-1}][1]}(x)$ for a time
        series $x$. A linear combination of sinusoidal waves with random Gaussian noise is used.}
    \label{fig:argmax}
\end{figure}

As an example, let $w=[1][1^2][23]$ and $x\in\tset[T]{3}$. For the arctic semiring, we now have
\begin{align*}
    \itsumu[\arctic]{w}(x)_T
        &=\max_{1\leq t_1\leq t_2\leq t_3\leq T}
            x_{t_1}^{\odot [1]}+x_{t_2}^{\odot [1^2]}+x_{t_3}^{\odot [23]}\\
        &=\max_{1\leq t_1\leq t_2\leq t_3\leq T}
            x_{t_1}^{[1]}+2x_{t_2}^{[1]}+x_{t_3}^{[2]}+x_{t_3}^{[3]}\\
        &=\max_{1\leq t_1\leq t_3\leq T}
            3x_{t_1}^{[1]}+x_{t_3}^{[2]}+x_{t_3}^{[3]}\\
        &=\itsumu[\arctic]{\tilde{w}}(x)_T,
\end{align*}
where $\tilde{w}=[1^3][23]$.
As observed in \cite{DEFT2022}, this phenomenon
leads to a ``collapse'' of most of the features for these modified iterated sums.
To circumvent this, we follow the suggestion in
\cite{DEFT2022}, and allow \emph{negative} exponents
${n_1,\dotsc,n_k\in\Z\setminus\{0\}}$ for an extended letter ${{a_i=[d_1^{n_1}\dotsc d_k^{n_k}]}}$.
In the arctic semiring, this leads to interesting objects.
If we for example use alternating signs ${w=[1^{1}][1^{-1}][1^{1}][1^{-1}]\dotsc}$, the transformation
$\itsumu[\arctic]{w}$ looks for a maximum, followed by the next minimum, followed by another
maximum, and so on.
The largest possible sum resulting from such constellations (where minima are multiplied by
$-1$) is returned. Figure~\ref{fig:argmax} shows an example of the indices of
maxima and minima for this arctic iterated sum for a word of length $3$,
$w=[1][1^{-1}][1]$. One can see that naively searching for those optima may take
time in $\mathcal{O}(T^3)$. Iterated sums can do this in linear time, as we will
see in Section~\ref{sec:compcomp}.

In the proceeding determination of the iterated sum,
the \emph{position} of the attained maxima/minima
is lost, but there is a way to simultaneously keep track of them,
\Cref{app:argmax_indices}.

\subsection{Computational complexity}\label{sec:compcomp}

While the definition in \eqref{eq:itsum} seems to involve the computation of $\mathcal{O}(T^p)$
terms for the entire sum, it is a well known fact that iterated sums (and integrals)
can be efficiently calculated using a simple dynamic programming principle.
This reduces the complexity to
$\mathcal{O}(T\cdot p)$
and is based on the following two facts, which follow from \Cref{thm:iss} in the appendix.

For a semiring $\semiring$,
let ${\cs[\semiring]{r}:\tset{1}\to\tset{1}},$
\begin{align*}
    \cs[\semiring]{r}(x)_s\coloneqq\left\{\begin{array}{ll}
        \zero&,1\leq s\leq r\\
        \bigoplus\limits_{j=1}^{s-r}x_j&,s>r
    \end{array}\right.,
\end{align*}
be the cumulative sum with an additional ``shift'' $r\in\N$.
\begin{fact}\label{thrm:cs_iss}
	For a word $w=[a_1]\cdots[a_p], p\geq 1$ and $x\in\tset{d}$
	\begin{align*}
		\itsum[\semiring]{w}(x)=\cs[\semiring]{0}\left(
            x^{[a_p]}\odot \cs[\semiring]{1}\left(
                x^{[a_{p-1}]}\odot  \cs[\semiring]{1}\big(\dotsc x^{[a_2]}
                \odot  \cs[\semiring]{1}(x^{[a_1]})\big)
            \right)\right),
	\end{align*}
	where ``$\odot$'' stands for the entry-wise product of time series in $\semiring$.
\end{fact}

When dealing with iterated sums that have non-strict index inequalities~\eqref{eq:itsum_semiring}, a slight modification can be made.
\begin{fact}\label{thrm:cs_iss_semiring}
	For a word $w=[a_1]\cdots[a_p], p\geq 1$ and $x\in\tset{d}$
	\begin{align}\label{eq:cs_iss_semiring}
        \itsumu[\semiring]{w}(x)=\cs[\semiring]{0}\left(x^{\odot [a_p]}\odot \cs[\semiring]{0}\left(
                x^{\odot [a_{p-1}]}\odot \cs[\semiring]{0}\big(\dotsc x^{\odot [a_2]}\odot
                    \cs[\semiring]{0}(x^{\odot [a_1]})\big)
            \right)\right),
	\end{align}
	where ``$\odot$'' stands for the entry-wise product of time series in $\semiring$.
\end{fact}

\subsection{Weighted iterated sums}\label{sec:weighting}

\newcommand\W{\omega}

Different weighting of information
at different time steps has been used in various applications,
e.g. in RNNs \cite{koutnik2014clockwork},
in transformers \cite{qin2021cosformer},
and recently, in the context of iterated sums \cite{Kri2021}
(we note related algebraic consideration in \cite[p.2]{foissy2021general}).

We see in \eqref{eq:iss_coeff} that all index combinations $t_1,\dotsc,t_p$ are treated equally.
In many applications, however, it is desirable to give more weight to combinations of indices that
are close to one another.
We introduce a weighting to $\itsum[\semiring]{w}$ that penalizes summands which are further apart
from one another.
We are thus looking
for a class of functions $\W:[T]^{p+1}\to S$, such that for the \DEF{weighted iterated sum}
\begin{align}\label{eq:weighted_iss}
    \itsum[\semiring,\W]{w}(x)_t :=\bigoplus_{0<t_1<\dotsc< t_p\leq t}
        \W(t_1,\dotsc,t_p,t)\odot x_{t_1}^{\odot [a_{1}]}\odot\cdots\odot x_{t_p}^{\odot [a_{p}]}.
\end{align}
\begin{enumerate}[label=\Roman*)]
    \item the calculation of the weighted iterated sum still is compatible with the dynamic
        programming procedure (\Cref{sec:compcomp}), and\label{enum:weighting1}
    \item $\W(t_1,\dotsc,t_p,t)$ is decreasing as $\abs{t_1-t}$ increases.
        \label{enum:weighting2}
\end{enumerate}
For the standard semiring $\reals$
an exponential weighting of the following form fulfills these requirements.
Let $g:\N\to\R$, be a non-decreasing function and
define
\begin{align*}
    \W^{\reals}_{\alpha_1,\dotsc,\alpha_{p}}(t_1,\dotsc,t_p,t)
        &\coloneqq e^{\alpha_1(g(t_1)-g(t_2))+\dotsc
            +\alpha_{p-1}(g(t_{p-1})-g(t_p))+\alpha_p(g(t_p)-g(t))}\\
        &=e^{\alpha_1g(t_1)}\cdot\left(\prod_{k=2}^{p}e^{(\alpha_k-\alpha_{k-1})g(t_k)}\right)
            \cdot e^{-\alpha_{p}g(t)}
\end{align*}
for $\alpha_1,\dotsc,\alpha_{p}\geq 0$ as a weighting on $\reals$.
For example, with $g(x) = x$,
$\alpha_1,\alpha_2,\alpha_3>0$,
and for a word of length
3, $w=[a_1][a_2][a_3]$ and $x\in\tset[T]{d}$
we obtain the weighted iterated sum
\begin{align*}
    \sum_{0<t_1<t_2<t_3\leq T}e^{\alpha_1(t_1-t_2)+\alpha_2(t_2-t_3)+\alpha_3(t_3-T)}
        x_{t_1}^{[a_1]}x_{t_2}^{[a_2]}x_{t_3}^{[a_3]}.
\end{align*}

Addition in the exponent becomes a multiplication of exponentials. Hence, with the definition
\begin{align}\label{eq:weighting_transform}
    y_{t_k} :=\exp((\alpha_{k}-\alpha_{k-1})g(t_k))x_{t_k}^{[a_k]}\text{ for }k=1,\dotsc,p
\end{align}
where $\alpha_0=0$, one can verify that the formula of Proposition~\ref{thrm:cs_iss}
still holds for the weighted iterated sum
\begin{align*}
    \itsum[\reals,\W^{\reals}_{\alpha_1,\dotsc,\alpha_{p}}]{w}(x)=\nu\cdot \cs[\reals]{0}\left(
        y^{[a_p]}\cdot\cs[\reals]{1}\left(
            y^{[a_{p-1}]}\cdot\cs[\reals]{1}\big(\dotsc y^{[a_2]}
            \cdot\cs[\reals]{1}(y^{[a_1]})\big)
        \right)\right)
\end{align*}
with $\nu=(e^{-\alpha_p\cdot 1},\dotsc,e^{-\alpha_p\cdot T})$.

For the arctic semiring, this argument does not work, as $\W^{\reals}$ does not fulfill
\ref{enum:weighting1}. We instead use
\begin{align*}
    \W_{\alpha_1,\dotsc,\alpha_{p}}^{\arctic}(t_1,\dotsc,t_p,t)
        &\coloneqq\alpha_1\cdot\Big(g(t_1)-g(t_2)\Big)+\dotsc+\alpha_{p}\cdot\Big(g(t_p)-g(t)\Big),
\end{align*}
which allows to repeat the argument above for
${y_{t_k}=(\alpha_{k}-\alpha_{k-1})g(t_k)+x_{t_k}^{\odot [a_k]}}$ and an appropriate ``additive'' $\nu$.

\bigskip

In either case, the function $g$ can actually depend on the time series $x$,
and we can choose it of the form
\begin{align}\label{eq:weighting_g}
    g(t)\coloneqq f(h(t, x)),\text{ where }f:[0,1]\to\R,\;h:\N\times\tset[T]{d}\to[0,1].
\end{align}
We experiment with different
$h$, e.g. the (normalized) sum of absolute increments\footnote{$\delta x_i\coloneqq x_i-x_{i-1}$; see also \Cref{sec:preparation}.}
\begin{align}\label{eq:weighting_h}
    h^{(\text{L1})}(t, x)\coloneqq \frac{\sum_{r=2}^{t}\abs{\delta x_r}}{\sum_{r'=2}^{T}\abs{\delta x_{r'}}}
\end{align}
or the sum of squared increments $h^{(\text{L2})}$. However, contrary to our initial
beliefs, setting $h^{\text{(id)}}(t,x)=\frac{t}{T}$ works best on the UCR archive.

The scaling function $f$ is chosen such that the weights have a non-vanishing impact in the
overall iterated sum. As the exponential often leads to exploding values, we restrict the range of
the weights to $[0,50]$, i.e. $f(x)=50x$.

In our experiments we will set $\alpha_1=\dotsc=\alpha_{p-1}=1$, so that the summands of the ISS
are just penalized over the total time range they cover,
as the sum of distances in the exponents becomes a telescoping sum.
We will use the shorter notation
$\itsum[\semiring,\W]{w}=\itsum[\semiring,\W^{\semiring}_{1,\dotsc,1}]{w}$.

\paragraph{Periodic Weightings}

\begin{figure}
    \includegraphics[width=\textwidth]{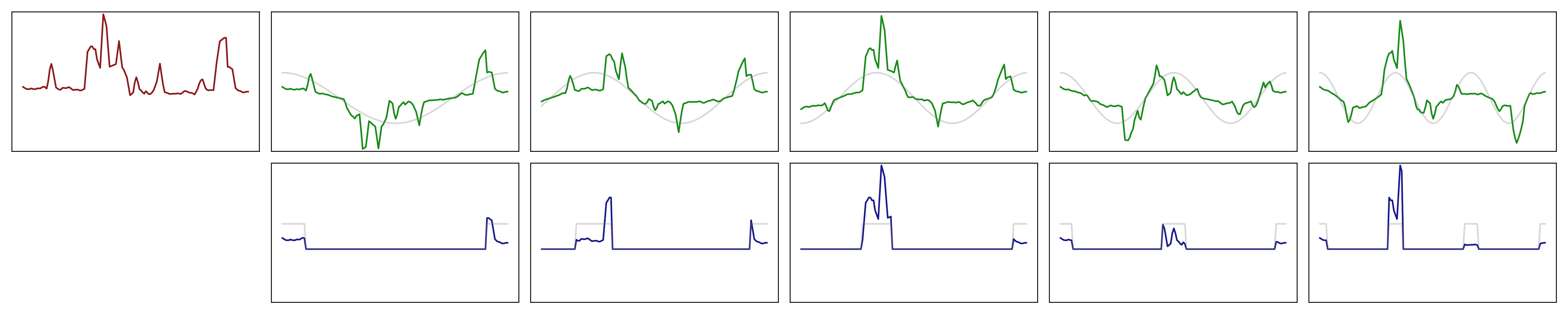}
    \caption{
        Example time series (red), multiplied with cosines/sines
        of different frequencies (green)
        and corresponding dilations
        of similar frequencies (blue).
        Multiplier in light gray.
        }
    \label{fig:powercons-cos}
\end{figure}

Allowing complex numbers $\gamma\in\C$ in the exponent $e^{\gamma(t_1-t_2)}$ allows for
periodical weightings. We explore cosine weightings, which is a special case of that form and is
easier to handle. In the standard semiring $\reals$, this weighting is defined by
\begin{align*}
    \W^{\cos^b}_f(t_1,\dotsc,t_p,t)=
        \prod_{k=2}^{p}\cos(\alpha_{k-1}(t_{k-1}-t_{k}))^b\cdot\cos(\alpha_p(t_p-t))^b,
\end{align*}
where $b\in\N$. We set the scalars $\alpha_1=\dotsc=\alpha_p=\frac{\pi}{f\cdot T}$ and $f\in[0,1]$
is a frequency parameter. For a word of length two and $b=1$, we can use the trigonometric
identity ${\cos(a-b)=\cos(a)\cos(b)+\sin(a)\sin(b)}$ in
\begin{align*}
    \itsum[{\W^{\cos}_f}]{[1][1]}(x)_t=
        &\sum_{0<t_1<t_2\leq t}
        &&\cos(\alpha_1(t_1-t_2))\cos(\alpha_2(t_2-t))x_{t_1}x_{t_2}\\
        =&\sum_{0<t_1<t_2\leq t}
        &&(\cos(\alpha_1t_1)\cos(\alpha_1t_2)+\sin(\alpha_1t_1)\sin(\alpha_1t_2))\\
        &&&\cdot(\cos(\alpha_2t_2)\cos(\alpha_2t)+\sin(\alpha_2t_2)\sin(\alpha_2t))x_{t_1}x_{t_2}\\
        =&\sum_{0<t_1<t_2\leq t}
        &&\cos(\alpha_1t_1)\cos(\alpha_1t_2)\cos(\alpha_2t_2)\cos(\alpha_2t)x_1x_2\\
        &&&+\cos(\alpha_1t_1)\cos(\alpha_1t_2)\sin(\alpha_2t_2)\sin(\alpha_2t)x_1x_2\\
        &&&+\sin(\alpha_1t_1)\sin(\alpha_1t_2)\cos(\alpha_2t_2)\cos(\alpha_2t)x_1x_2\\
        &&&+\sin(\alpha_1t_1)\sin(\alpha_1t_2)\sin(\alpha_2t_2)\sin(\alpha_2t)x_1x_2,
\end{align*}
to arrive again at an expression suited for dynamic programming.
Note that this last expression is the sum of
four iterated sums. It is straight-forward to write in a similar fashion,
an algorithm for words of any length and arbitrary $b\in\N$.
Cosine weighting can be thought of as a continuous version of a spacing, or dilation, operation,
\Cref{fig:powercons-cos}.

\section{FRUITS pipeline}
\label{sec:pipeline}

The \algoname{Fruits} pipeline comprises three steps. Each one has its own set of
hyperparameters.
An entire pipeline is defined by a configuration of these hyperparameters,
and we will refer to one such configuration as a ``fruit''. We restrict the following
discussion to the most vital parts of the pipeline. We make our code available as a Python
package\footnote{\url{https://github.com/irkri/fruits}}. The package is written in an
object-oriented programming style, which makes it easy to use and customize.

\begin{figure}
    \centering
    \includegraphics[width=.8\textwidth]{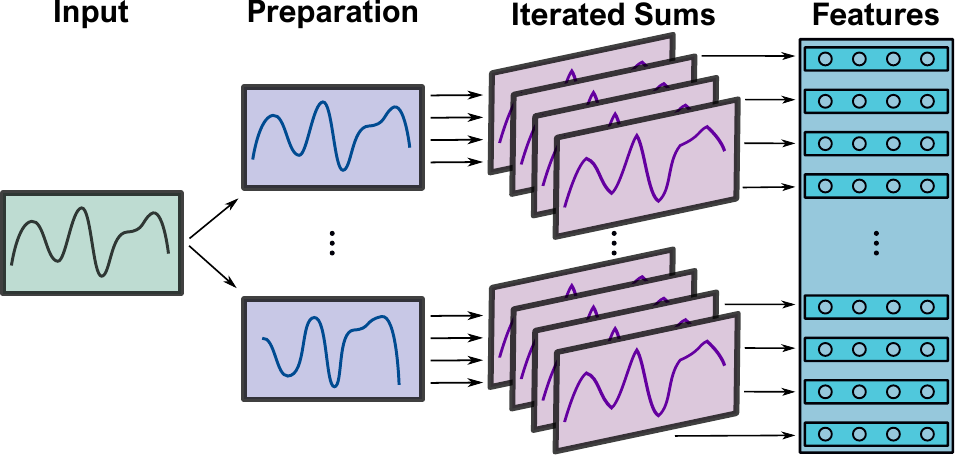}
    \caption{The \algoname{Fruits} pipeline.}
    \label{fig:pipeline}
\end{figure}

The following subsections will introduce transformations in each of the three steps of the
\algoname{Fruits} pipeline, \Cref{fig:pipeline}. First, \algoname{Fruits} will preprocess the input data in the
``preparation step''. These transformations, introduced in Section~\ref{sec:preparation} will be called
\textit{preparateurs}. After preparation, the iterated sums for different semirings, words and
weighting configurations are calculated. The result of this second step is a (family of) transformed
one-dimensional time series.
The last step then extract features from these time series, a process we
call \textit{sieving}. The transformations here will be called \textit{feature sieves}.

\subsection{Preparation}\label{sec:preparation}

The data preparation stage preprocesses the input data before the calculation of
the iterated sums.

\paragraph{Standardization}

Standardization, a very common transform for machine learning pipelines,
has a significant impact on the accuracy performance of \algoname{Fruits}.
We apply it in each of the pipelines at the very beginning,
normalizing every single time series to have mean zero and standard deviation one.
We also experimented with other normalizations, e.g. to the interval $[0,1]$,
but found standardization to work best.

\paragraph{Increments}
The increments of a time series,
\begin{align*}
    \finc:\tset{d}\to\tset{d},x\mapsto(0,x_2-x_1,x_3-x_2,\dotsc)\eqqcolon \delta x,
\end{align*}
can be considered the discrete analog of a derivative.
Calculating the ISS on it leads to \textit{time-warping invariant} features $\itsum{w}(\delta x)_T$, see
\cite{DEFT2020}, \cite{DEFT2022} for an in-depth discussion. For example, two time series
\begin{align*}
    x=(x_1,x_2,&x_3,x_4,x_5,x_6,x_7,\dotsc,x_{T-2},x_{T-1},x_T)\in\tset[T]{d},\\
    y=(x_1,x_2,&x_3,x_3,x_3,x_4,x_5,x_6,x_7,\dotsc,x_{T-2},x_{T-1},x_T)\in\tset[T+2]{d}
\end{align*}
have the same feature representation using this preparateur and appropriate feature sieves. We will
discuss details in Section~\ref{sec:sieving}. Here we say that $y$ is the time series obtained
by \textit{stuttering} $x$ at time step $3$ two times.

Experiments showed that lifting a univariate time series to two dimensions,
where the second dimension is given by its increments
leads to a significant improvement in accuracy. We denote this preparateur map by
\begin{align}
    \label{eq:fincel}
    \begin{split}
        \fincel:\tset{1}&\to\tset{2}\\
        x=(x_1,x_2,x_3,\dotsc)&\mapsto \left(
            \begin{pmatrix}x_1\\0\end{pmatrix},\begin{pmatrix}x_2\\x_2-x_1\end{pmatrix},
            \begin{pmatrix}x_3\\x_3-x_2\end{pmatrix},\dotsc
        \right).
    \end{split}
\end{align}

Further down the pipeline, using words over an alphabet with two dimensions $A=\{1,2\}$, we get interesting expressions like
\begin{align*}
    \itsum{[12]}(\fincel(x))_T=\sum_{t\leq T} x_i(x_i-x_{t-1})
        =\sum_{t\leq T} x_t^2-x_tx_{t-1}=\itsum{[11]}(x)_T-\sum_{t\leq T}x_tx_{t-1}.
\end{align*}
The second term cannot be easily approximated by linear transformations on iterated sums of just
$x$. However, we want to note that it can be approximated by choosing an
appropriate weighting control function $g$, see \eqref{eq:weighting_g}.
We only need to ensure $e^{g(t_1)-g(t_2)}\approx 0$ for $\abs{t_1-t_2}>1$.

\subsection{Iterated sums}

The central part of \algoname{Fruits}, the iterated-sums signature ISS, is applied to the
preprocessed time series, obtained from one or multiple preparateurs presented in
Section~\ref{sec:preparation}.
Several options can be considered for the ISS.
\begin{enumerate}
    \item The words the ISS is calculated for.
    \item The semiring in which the operations $\oplus,\odot$ are performed.
    \item The type of weighting we apply to the signature.
\end{enumerate}
Experiments suggesting the best options are presented in Section~\ref{sec:experiments}.

\subsection{Feature sieving}\label{sec:sieving}

The output of $\itsum{w}(x)$ is a one-dimensional time series $z\in\tset[T]{1}$ of the same length
$T$ as the input $x\in\tset[T]{d}$. Using a preselected set of words $w$, we get a number of
iterated sums for each sample in the dataset. We will now extract representative features from the
iterated sums, a process we call \textit{sieving}. As the calculation of iterated sums for a large number of words
is expensive, extracting several meaningful features from each iterated sum reduces
computation costs and increases feature variety.

\paragraph{Last value}
A natural feature for $x\in\tset[T]{d}$ is the total iterated sum $\itsum[\semiring]{w}(x)_T$, the
corresponding feature sieve is named $\fend: z\mapsto z_T$. The entire input time series is
considered for this feature.

\paragraph{Coquantiles}

We can also cut the time series at some time $s<T$, i.e. considering a sieve $z\mapsto z_s$. To
again get time-warping invariant features, we introduce \textit{coquantiles}. For a given $0<q<1$
we define
\begin{align*}
    \zeta^G_q(x)\coloneqq\max\left\{
        t\in\{1,\dotsc,T\}\mid \big(G(\delta x)\big)_t
        \leq q\cdot\sup\big(G(\delta x)\big)
    \right\}
\end{align*}
for a function $G:\tset[T]{d}\to\tset[T]{1}$ returning monotonic increasing time series. We set
$\zeta^w_0(x)\coloneqq 1$ and $\zeta^w_1(x)\coloneqq T$. A natural choice is
\begin{align*}
    \zeta^{h^{(\text{L1})}}_q(x)=
        \max\left\{t\in\{1,\dotsc,T\}\mid h^{(\text{L1})}(t)\leq q\right\}
\end{align*}
where $G=h^{(\text{L1})}$, see \eqref{eq:weighting_h}. For this choice,
\begin{align*}
   \itsum[\semiring]{w}(y)_{\zeta^G_q},
\end{align*}
is invariant under stuttering. However, experiments on the UCR archive show that coquantiles do not
lead to a better performance of our pipelines. We will omit them in later discussions and leave
this paragraph as a remark.

\paragraph{Number of positive increments}
\citeauthor{DPW2020} \cite{DPW2020} found that for \algoname{Rocket}, the feature
\textit{Proportion of Positive Values (PPV)} is essential for their good performance.
Loosely speaking, this transform calculates the relative number of occurrences of a certain pattern in
a time series. In the case of \algoname{Rocket}, this pattern is given by a (randomized) kernel.
The performance of PPV highly depends on a good choice of a reference value, for which numbers greater than this value are
considered ``positive''. In \algoname{Rocket}, this reference value is a quantile of a convolution
of one sample in the training dataset.

Inspired by this, \algoname{Fruits} calculates the \textit{Number of Positive Increments}
$\fnpi:\tset{1}\to [0,1]$,
\begin{align*}
    \fnpi(z)\coloneqq \sum_{t=1}^T \mathds{1}_{\finc(z)_t>0}\text{ where }\mathds{1}_{a>0}=\left\{
    \begin{array}{cc}
        0, &a\leq 0\\
        1, &a > 0
    \end{array}\right.,
\end{align*}
which is the number of time steps for which the $z$ is increasing. We note that this sieve is
automatically time-warping invariant. One immediate consequence of the definition is that we get
constant values $\fnpi(z)\in\mathcal{O}(T)$ for strictly monotonically increasing iterated sums
$z=\itsum[\semiring]{w}(y)$, e.g. for ${w=[11]}, {w=[1111]}, {w=[11][11]}$ (we write $\mathcal{O}(T)$
as the actual result depends on the length of $w$). We generalize $\fnpi$ to $k$-th order
increments $\finc^k(z)=\finc(\finc(\dotsc\finc(z)))$ ($k$-times) and write
\begin{align*}
    \fnpi^k(z)\coloneqq \sum_{t=1}^T \mathds{1}_{\finc^k(z)_t>0}
\end{align*}
with the special case $\fnpi^0(z)=T\cdot\operatorname{PPV}(z)$. Additionally, we restrict $\fnpi$
to only count positive increments in a certain window
\begin{align*}
    \fnpi^k_{\alpha_l,\alpha_r}(z)\coloneqq \sum_{t=1}^T \mathds{1}_{q_l<\finc^k(z)_t\leq q_r}.
\end{align*}
This window is given by quantiles $q_l$ and $q_r$, which are the estimated $\alpha_l$- and
$\alpha_r$-quantiles of the iterated sums of samples from the training set. In most experiments we
set $\alpha_l=0.5$ and $\alpha_r=1$, for which we define $q_r=\infty$.

Using the arctic semiring $\semiring=\arctic$, $\fnpi(\itsum[\arctic]{w}(y))$ counts the number of
times the outer maximum of the iterated sum
\begin{align*}
    \max_{1\leq t_1\leq\dotsc\leq t_p\leq T}y_{t_1}^{\odot [a_1]}+\dotsc+y_{t_p}^{\odot [a_p]}
    \text{ for }w=[a_1]\dotsc[a_p]
\end{align*}
changes. $\fnpi^2(\itsum[\arctic]{w}(y))$ on the other hand has a less intuitive interpretation. It
counts how often small changes are followed by larger changes of this maximum.
Experiments show that this feature sieve, for both the standard as well as the
arctic semiring boost the performance of the pipeline.

\paragraph{Mean of positive increments}
\algoname{MultiRocket} \cite{DSW2021} also introduces variations of $\operatorname{PPV}$ to the
\algoname{Rocket} pipeline. We found that one such variation also improves the performance of
\algoname{Fruits}. Similarly to $\fnpi$, we change the $\operatorname{MPV}$ operation from
\algoname{MultiRocket} to
\begin{align*}
    \fmpi^k_{\alpha_l,\alpha_r}(z)\coloneqq\frac{1}{T}\sum_{t=1}^T
        \mathds{1}_{q_l<\finc^k(z)_t\leq q_r}\finc^k(z)_t,
\end{align*}
which is the mean of positive increments in a window restricted by estimated quantiles.

\section{Experiments}\label{sec:experiments}

We conduct extensive experiments on the datasets in the UCR-archive \cite{Dau2019}.
We will use a linear regression classifier with L2 regularization in all experiments,
the same classifier used in \algoname{Rocket} \cite{DPW2020}. Using cross-validation, a suitable
regularization parameter can be found quickly for this ridge regression.

\subsection{Results on specific datasets}\label{sec:exp_datasets}

This section presents novel observations we made on several datasets from the UCR
archive, some of which motivate certain choices we made for the \algoname{Fruits} pipeline.

\paragraph{ChlorineConcentration}
In contrast to most datasets in the UCR-archive, ChlorineConcentration (see \cite{LMPF2009}) seems
to be mostly unaffected by changes to the \algoname{Fruits} pipeline and performs badly compared to
\algoname{Rocket}.

The dataset comprises 4310 time series, 467 of which are for training and 3840 are for testing. The
time series length is 166. A software called EPANET was used to simulate a water distribution
piping system where the concentration of chlorine in the water was measured at 166 nodes within the
system. Each 5 minutes in a duration of 15 days, one such measurement was taken. This yields a
total of 4310 time steps. This description is contradictory to the dataset structure. The temporal
structure of a sample is given by measurements on different nodes across the piping system and not
by the 4310 time steps. We believe that this labeling is responsible for the bad performance of
\algoname{Fruits}.

\begin{figure}
    \begin{minipage}{0.33\textwidth}
        \includegraphics[width=\textwidth]{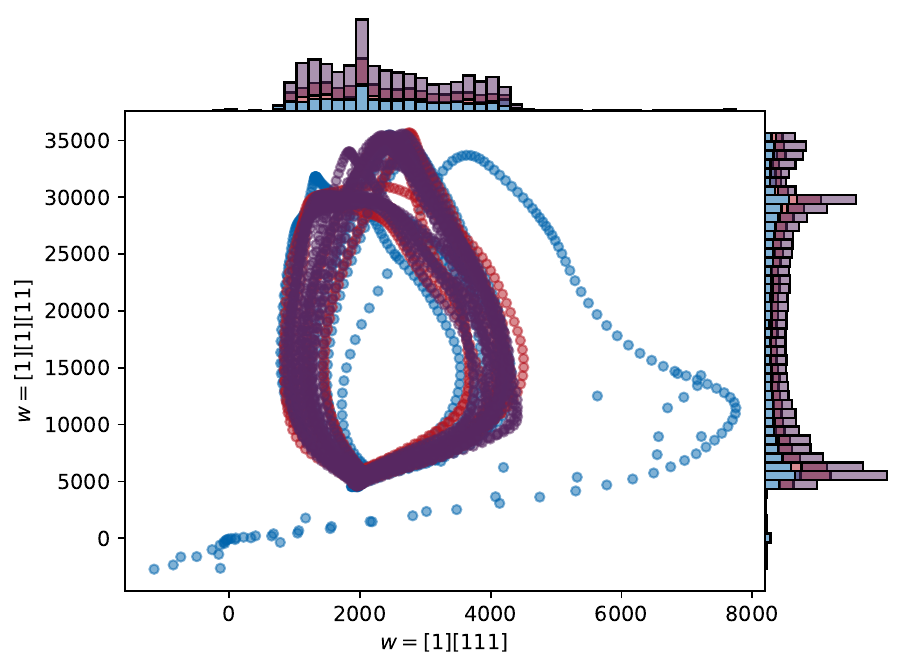}
    \end{minipage}\hfill
    \begin{minipage}{0.33\textwidth}
        \includegraphics[width=\textwidth]{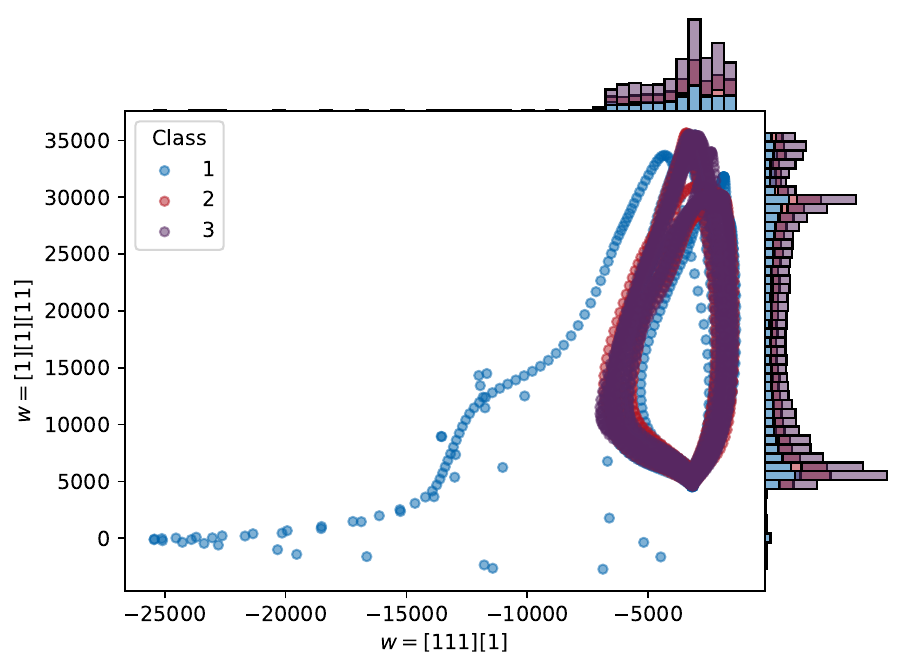}
    \end{minipage}\hfill
    \begin{minipage}{0.33\textwidth}
        \includegraphics[width=\textwidth]{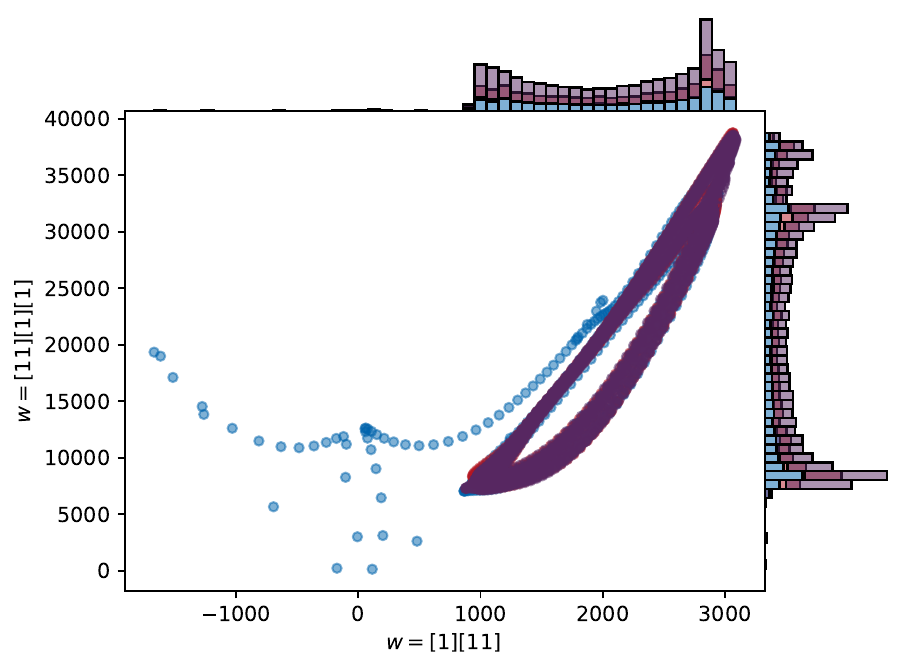}
    \end{minipage}
    \caption{Scatter plot of different \algoname{Fruits} features $\itsum{w}(x)_T$ for different
    words $w$, where $x$ is a time series from the dataset ChlorineConcentration. The plots also
    show histograms of the feature distribution of each single word.}
    \label{fig:chlorineconcentration}
\end{figure}

Plotting the features returned by a simple \algoname{Fruits} configuration in a scatter
plot\footnote{All scatter plots and critical difference diagrams were made using the code publicly
available at \url{https://github.com/irkri/classically}.} results in interesting topological
structures, see Figure~\ref{fig:chlorineconcentration}. The classes in the dataset are recorded
time series for different concentrations of chlorine in the water. The classes 2 and 3 are
distributed on ring-like structures while the first class seems to be more scattered.

\paragraph{GesturePebbleZ1/2}
The datasets GesturePebbleZ1 and GesturePebbleZ2 motivate the replacement of missing values in the
time series with the last observed value. We compare two settings: replacing NaN (not-a-number)
entries in the time series by $0$ and replacing them by the last recorded (non-NaN) value. In both
cases we use features $\itsum[\reals,W]{w}(\finc(x))_N$ for words $w$ of weight $\leq 4$ and the
weighting control function $h^{(\text{L1})}$, see \eqref{eq:weighting_h}. If NaN values appear at
the start of the time series, the last recorded value is defined to be $0$. The first setting
produced an accuracy of $15.82\%$ while the second one yields $29.75\%$. If we do not use the
$\finc$ preparateur, the accuracies are $15.82\%$ and $48.10\%$ respectively.

\paragraph{PigCVP/PigArtPressure/PigAirwayPressure} These datasets are presented in \cite{GBD2017},
where the authors had the goal to classify a three-dimensional time series as being vital signals
from different pigs before or after an internal bleeding occurred. In the UCR archive, the
dimensions are split into three independent datasets and the classification task is now to predict
from which of the 52 different pigs one sample originates. For that, one time series is cut in
three equivalently sized parts, one for the training set and two for the test set. As there are two
time series for each pig, before and after internal bleeding, the datasets have $104$ training
samples and $208$ test samples in the UCR version. The starting points of the time series at $t=1$
in the training and test set therefore have a different meaning. Again, this dataset surfaced in
our experiments as different \algoname{Fruits} configurations did not perform well, in contrast to
\algoname{Rocket}. Intuitively, finding time series in the test set that extend the ones in the
training set might be more of a pattern-matching task, which \algoname{Rocket} essentially is
suited best for.

\paragraph{Yoga}
This dataset of image-derived time series records distances of a persons outline to its body center
while the person is doing different yoga poses. The goal here is to distinguish between a male and
female person doing the same poses. Here, using just $\fnpi(\itsum[\reals]{[1^4]}(\finc(x)))$ as a
feature in a linear classifier results in an accuracy of $85.37\%$. Another fruit calculating 300
features using words of weight 1-4 and $\fend$ with ten different coquantiles on both $x$ and
$\finc(x)$ only achieves $69.77\%$.

\subsection{General-Purpose Fruit}\label{sec:general_purpose_fruit}

The following subsections will outline our process of finding a collection of \algoname{Fruits}
components with their corresponding best working configuration. We decide this based on accuracy on
the UCR archive as well as the compute time needed by these pipelines.

The resulting pipeline is described by the following three sequences of components, whose output
is put into a linear Ridge regression classifier.
\begin{itemize}[label=$\bullet$]
    \item $\fincel\to\operatorname{STD}\longrightarrow
        \bigl\{\itsum[\reals,\W]{w}\mid \abs{w}\leq 6\bigr\}
        \longrightarrow \mathcal{S}$
    \item $\fincel\longrightarrow
        \bigl\{\itsum[\arctic]{w}\mid
            w=\underbrace{[a^{\pm 1}][b^{\mp 1}][a^{\pm 1}]\dotsc}_{\text{length }48},
            a,b\in\{1,2\}\bigr\}
        \longrightarrow\mathcal{S}$
    \item $\fincel\to\operatorname{STD}\longrightarrow
        \bigl\{\itsum[{\W^{\cos^1}_f}]{w},\itsum[{\W^{\cos^2}_f}]{w}\mid
        \abs{w}\leq 4,f\in\{i/20\mid i=1,3,5,7,9\}\bigr\}\longrightarrow \mathcal{S}$
\end{itemize}
For the set of sieves, we use for each of the three ISS
\[\mathcal{S}=\{
    \fnpi^0_{\frac{1}{2},1},\fnpi^1_{\frac{1}{2},1},\fnpi^2_{\frac{1}{2},1},
    \fmpi^0_{\frac{1}{2},1},\fmpi^1_{\frac{1}{2},1},\fmpi^2_{\frac{1}{2},1},\fend
\}.\]

\subsubsection{Preparation}

\begin{figure}
    \begin{minipage}{0.49\textwidth}
        \includegraphics[width=1.0\textwidth]{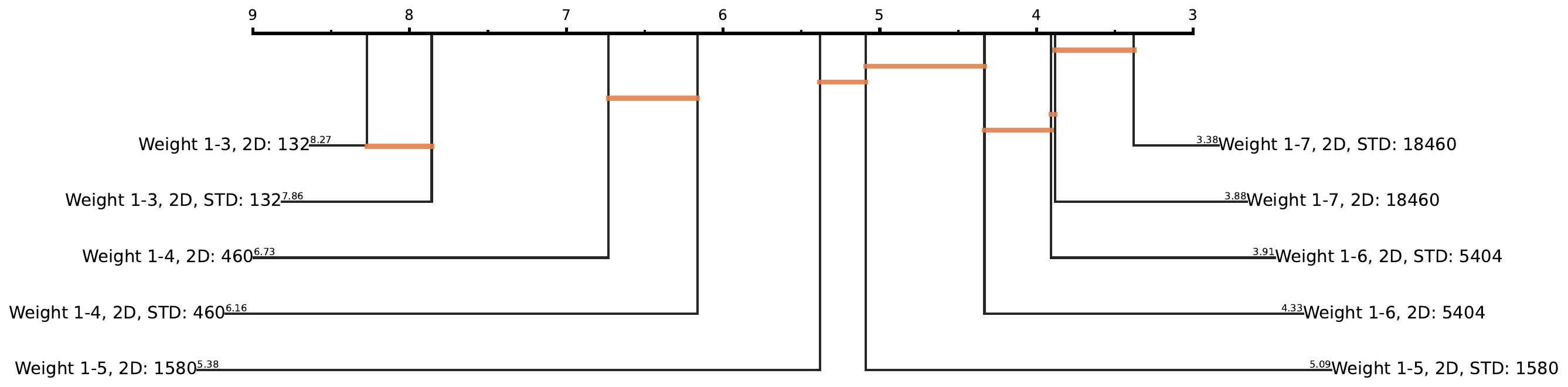}
        \caption{Critical difference diagram of fruits with and without standardization for
            increasing maximum word weight. Each label of a fruit also contains the total number of
            features used in this configuration. For example, the fruit with weight 1-5 yields 1580
            features.}
        \label{fig:words_weight_std}
    \end{minipage}\hfill
    \begin{minipage}{0.49\textwidth}
        \includegraphics[width=1.0\textwidth]{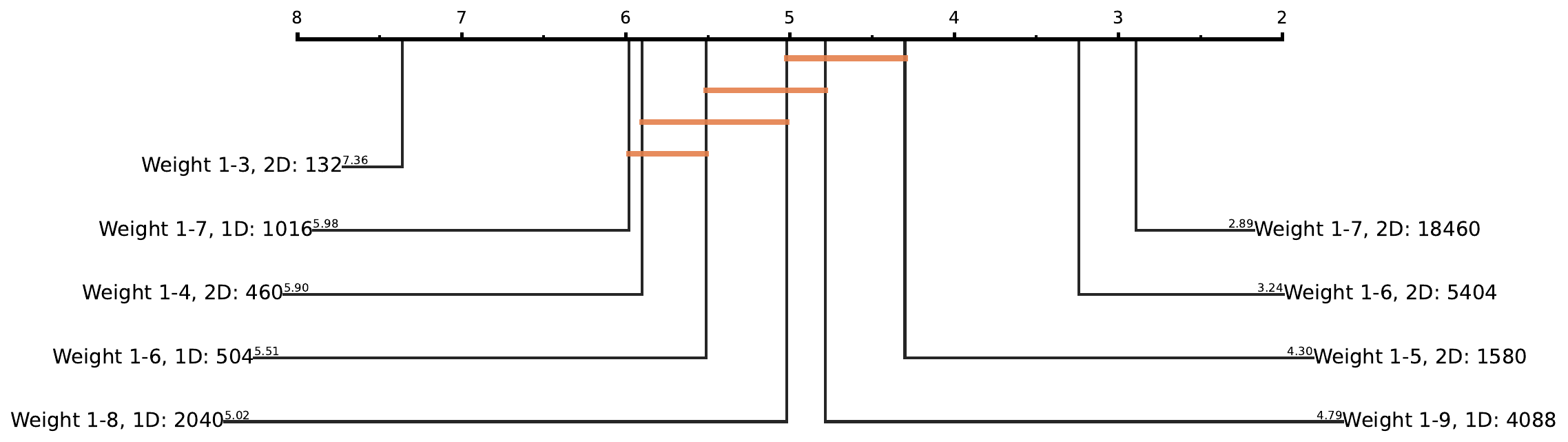}
        \caption{Comparison of fruits using a one or two-dimensional time series as input. The one
            dimensional version has the original time series and its increments processed
            separately. The two-dimensional version uses $\fincel(x)$.}
        \label{fig:words_weight_dim}
    \end{minipage}
\end{figure}

We examine the impact of standardizing input time series on
the performance of various \emph{fruits}, our name for complete pipelines.
As illustrated in Figure~\ref{fig:words_weight_std}, we employ a critical
difference diagram, following \cite{Dem2006}, to rank and compare the
accuracy of each fruit configuration across all datasets in the UCR archive.
In this diagram, configurations are ranked based on their average accuracy,
with lower ranks indicating higher accuracy. To assess the statistical
significance of the differences in performance, it uses two-sided Wilcoxon
signed-rank tests, applying a Bonferroni-Holm correction to account for
multiple comparisons. The diagram visually represents non-significant
differences by connecting the respective fruits with a line. It is important
to note that the expressiveness of these lines decreases as the number of
compared configurations increases, due to the adjusted significance levels
required by the Bonferroni-Holm correction.

Figure~\ref{fig:words_weight_std} shows that a standardized input leads to better results in all
configurations. This might not be a significant difference, but it also closes the gap between the
next higher total weight of words used. For example, a fruit with words of weight up to 6 are with
prior standardization not significantly different to a configuration with weight 7.

Figure~\ref{fig:words_weight_dim} compares the influence of the $\fincel$ transform
\eqref{eq:fincel} to the one dimensional version, where the input time series and its increments
are processed separately. In two dimensions, words like $[1][2]$ can be used to mix the two
signals. The plot shows that words of maximum weight 5 in two dimensions surpass a configuration
with maximum weight 9 in one dimension, while also being considerably smaller in feature size.

\subsubsection{ISS Configurations}

\paragraph{Weighting}

\begin{figure}
    \begin{minipage}{0.49\textwidth}
        \includegraphics[width=1.0\textwidth]{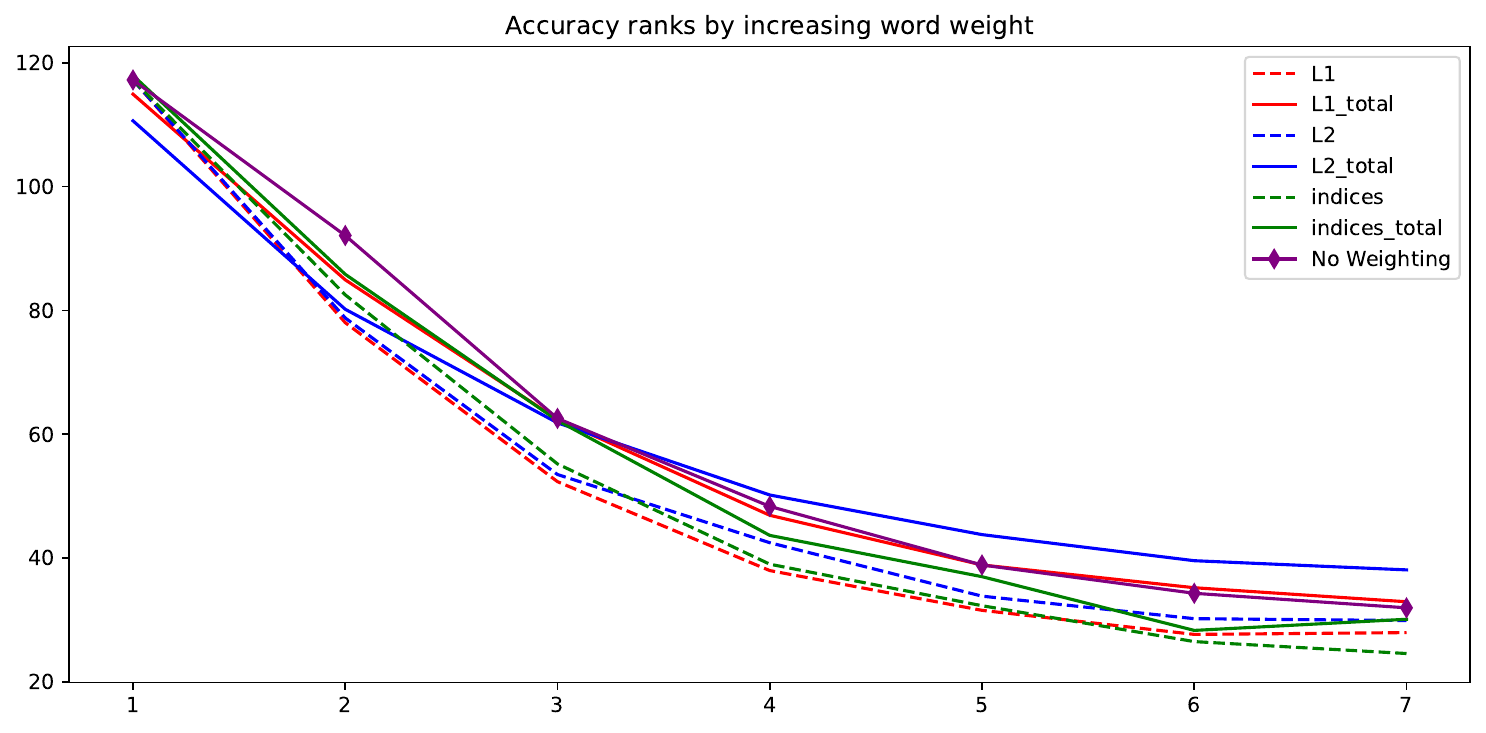}
        \caption{Ranks for the standard semiring $\reals$.}
        \label{fig:avg_ranks_weightings_reals}
    \end{minipage}\hfill
    \begin{minipage}{0.49\textwidth}
        \includegraphics[width=1.0\textwidth]{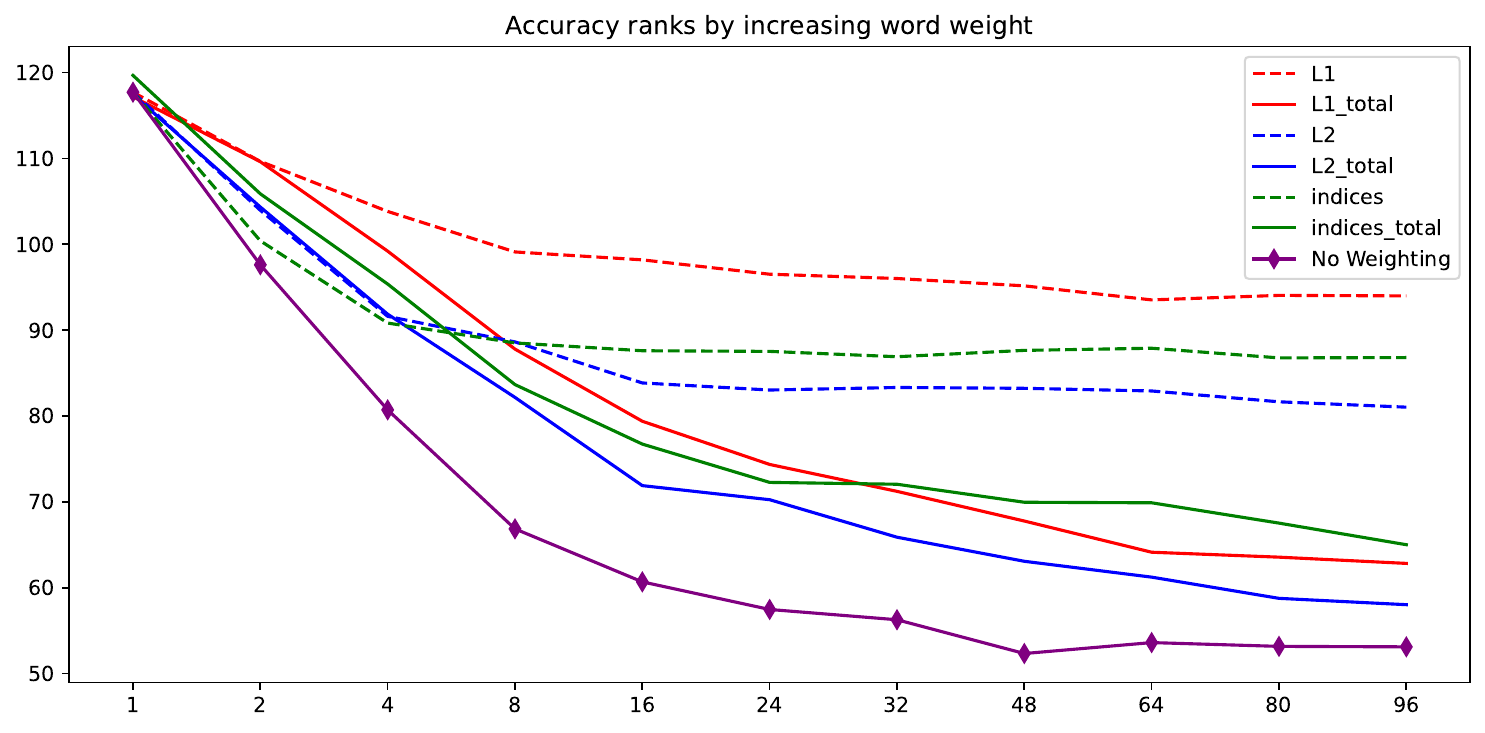}
        \caption{Ranks for the arctic semiring $\arctic$.}
        \label{fig:avg_ranks_weightings_arctic}
    \end{minipage}
    \caption*{Average ranks of fruits with different ISS weightings for increasing word weight on
        the UCR archive (lower rank is better). Labels are for the corresponding weighting $h$ and
        an additional ``\_total'' is appended if the weighting is for $\alpha_p\neq 0$.}
\end{figure}

\begin{figure}
    \begin{minipage}{0.3\textwidth}
        \includegraphics[width=1.0\textwidth]{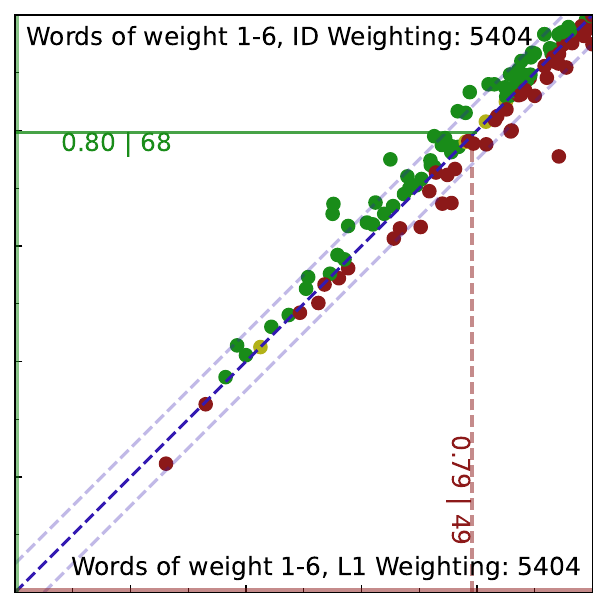}
        \caption{Comparison of $h^{\text{(id)}}$ and $h^{\text{(L1)}}$ in $\reals$ with words of
            weight one to six.}
        \label{fig:comp_reals_L1_ID}
    \end{minipage}\hfill
    \begin{minipage}{0.65\textwidth}
        \includegraphics[width=1.0\textwidth]{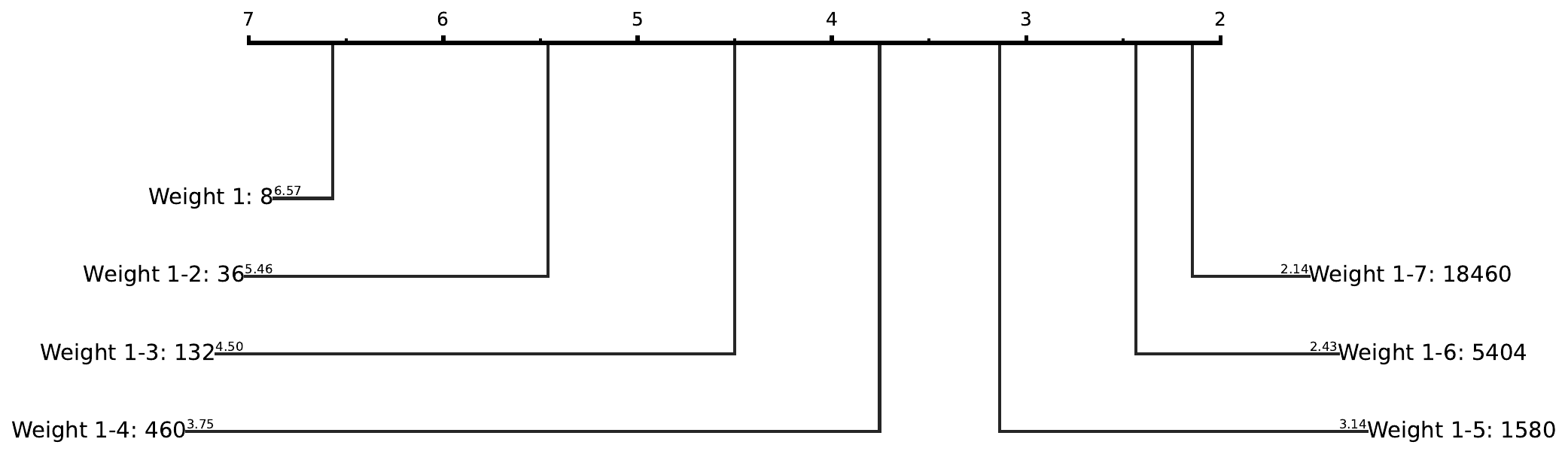}
        \caption{Critical difference diagram of different word weights for the weighting function
            $h^{\text(id)}$.}
        \label{fig:comp_reals_ID_word_weights}
    \end{minipage}
\end{figure}

Section~\ref{sec:weighting} introduces different weightings on the ISS. An open question is what
weighting function $h$ \eqref{eq:weighting_g} to pick. We set the scaling function $f(x)=50x$ and
compare $h^{\text{(L1)}},h^{\text{(L2)}},h^{\text{(id)}}$ for different weights of words in
Figure~\ref{fig:avg_ranks_weightings_reals}. Additionally, we experiment if the weighting for the
total iterated sum $e^{t_p-T}$ has an influence in accuracy. It turns out that for $\reals$, we get
better results without this outer weighting, i.e. setting $\alpha_p=0$ in
\eqref{eq:weighting_transform}. We see that in the standard semiring $\reals$, the L1 penalization
seems like the best choice for accuracy. However, as it is very close to the performance of just
the indices, i.e. $h^{\text{(id)}}$, this seems to be a more natural and computational efficient
choice. In the arctic semiring $\arctic$ (Figure~\ref{fig:avg_ranks_weightings_arctic}), no
weighting seems to work best. We have to note that these plots may not be the perfect way of
deciding the best configurations, as they only allow comparing the average rank of classifiers over
the whole UCR archive. Zooming in on one such comparison, namely $h^{\text{(id)}}$ and
$h^{\text{(L1)}}$, Figure~\ref{fig:comp_reals_L1_ID} reveals the detailed differences, where
$h^{\text{(id)}}$ surpasses $h^{\text{(L1)}}$ on 68 out of 128 datasets.

\paragraph{Words}

Figure~\ref{fig:avg_ranks_weightings_reals}~and~\ref{fig:avg_ranks_weightings_arctic} both also
help us to decide at which word weight to cut the ISS off. For $\reals$, weight $6$ seems like a
good candidate as the rank is not changing much to weight $7$. This is also shown in the critical
difference diagram of the word weights, see Figure~\ref{fig:comp_reals_ID_word_weights}. Each
weight actually increases the accuracies significantly, but the ranks of 6 and 7 are much closer
together. The time needed to compute weight 7 does not justify its use for longer time series. In
the arctic semiring $\arctic$ we will use words with alternating exponents
$[1^{+1}][1^{-1}][1^{+1}]\dotsc$ and $[1^{-1}][1^{+1}][1^{-1}]\dotsc$ up to length 48. All of these
experiments are made on two-dimensional time series $\fincel(x)$, so we actually use the
alternating counterparts of the words $[1][1][1]\dotsc$, $[2][2][2]\dotsc$ as well as
$[1][2][1]\dotsc$ and $[2][1][2]\dotsc$ for mixing the input time series and its increments.

\paragraph{Cosine Weighting}

Our standard configuration for a cosine weighted ISS is using words up to weight 4 (with
$\alpha_1=\dotsc=\alpha_p=1$) and 10 frequencies $f$ (compare Section~\ref{sec:weighting}). For a
single cosine weighted iterated sum of length 3, we have to compute 9 iterated sums. Increasing the
number of frequencies therefore is a large increase in compute time needed.

\begin{figure}
    \includegraphics[width=1.0\textwidth]{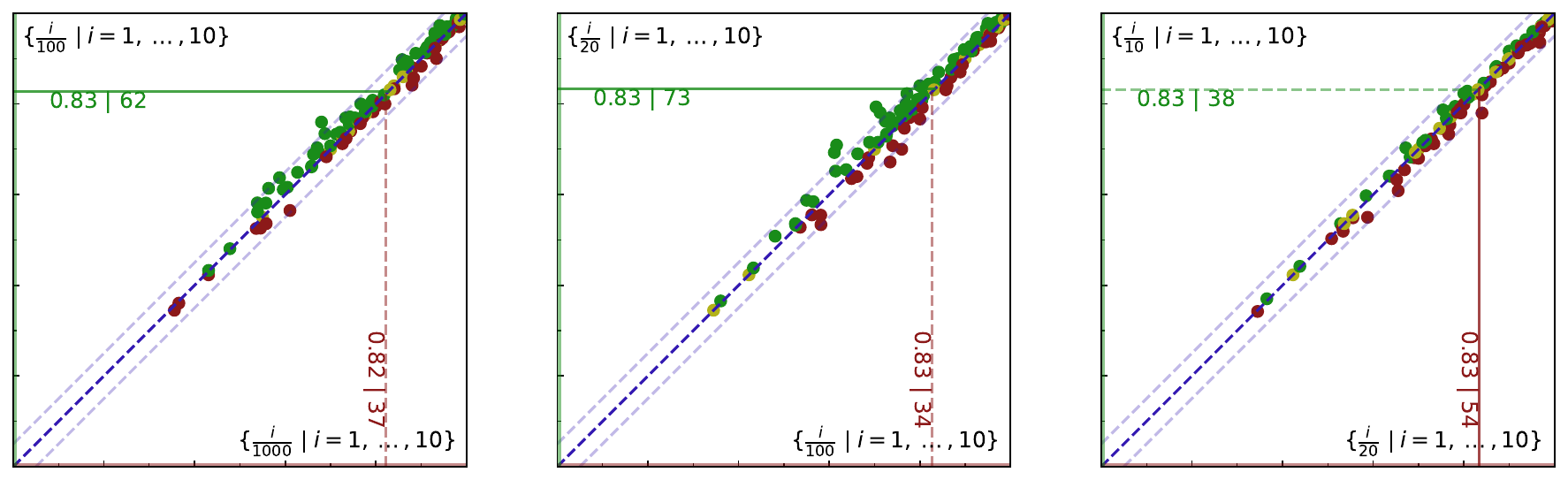}
    \caption{Comparison of four possible sets $\mathcal{F}\subset(0,1]$ for the frequency parameter
        $f\in\mathcal{F}$.}
    \label{fig:frequency_sets}
\end{figure}

Not only is the number of frequencies used important, but also their range. A frequency parameter
$f=0.9$ corresponds to a ``kernel spacing'' of 90\% of the time series length.
Figure~\ref{fig:frequency_sets} shows that the set $\mathcal{F}=\left\{\frac{i}{20}\mid
i=1,\dotsc,10\right\}$ gives the best overall results on the UCR archive out of the four tested. It
performs better than also using values $f>0.5$.

With this set $\mathcal{F}$, we can further reduce the number of frequencies to 5
($\left\{\frac{i}{20}\mid i=1,3,5,7,9\right\}$). The mean of absolute differences in accuracy of
the two pipelines is $\approx 1.1\%$, while using 5 frequencies being twice as fast as using 10. In
practice, this can mean a speed-up of about $2000$ seconds on large datasets in the UCR archive.

\subsubsection{Feature Sieve Selection}

We presented three feature sieves in Section~\ref{sec:sieving}. As a default, we will use for all
pipelines in total seven variations of them. $\fnpi$ and $\fmpi$ will use quantiles for
$\alpha_l=0.5,\alpha_r=1.0$ and will be evaluated on the pure iterated sums, its increments, and
its second order increments. The final sieves are
\begin{align*}
    \fnpi^0_{\frac{1}{2},1},\fnpi^1_{\frac{1}{2},1},\fnpi^2_{\frac{1}{2},1},
    \fmpi^0_{\frac{1}{2},1},\fmpi^1_{\frac{1}{2},1},\fmpi^2_{\frac{1}{2},1},\fend.
\end{align*}

\begin{figure}
    \begin{minipage}{0.49\textwidth}
        \includegraphics[width=1.0\textwidth]{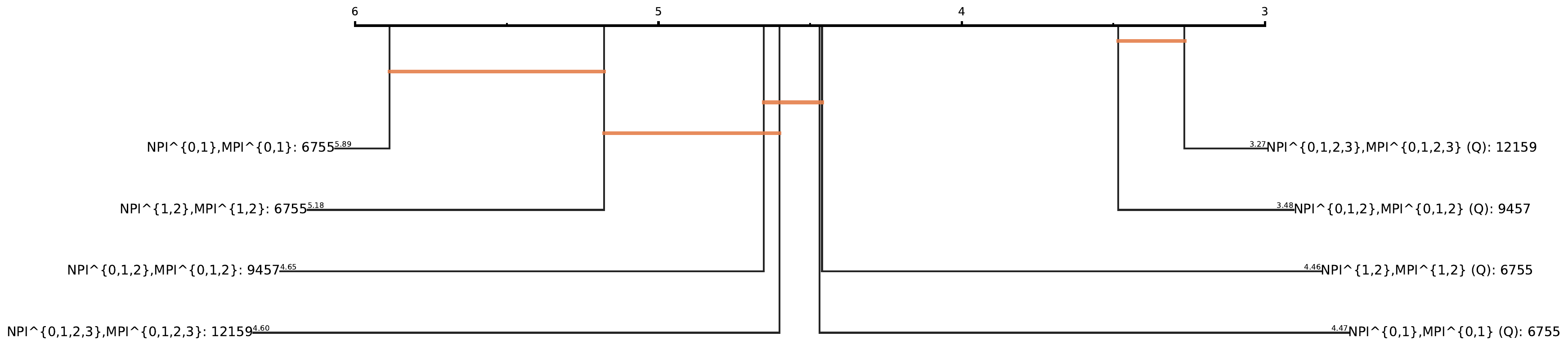}
        \caption{Standard semiring $\reals$}
        \label{fig:reals_sieves_cd}
    \end{minipage}\hfill
    \begin{minipage}{0.49\textwidth}
        \includegraphics[width=1.0\textwidth]{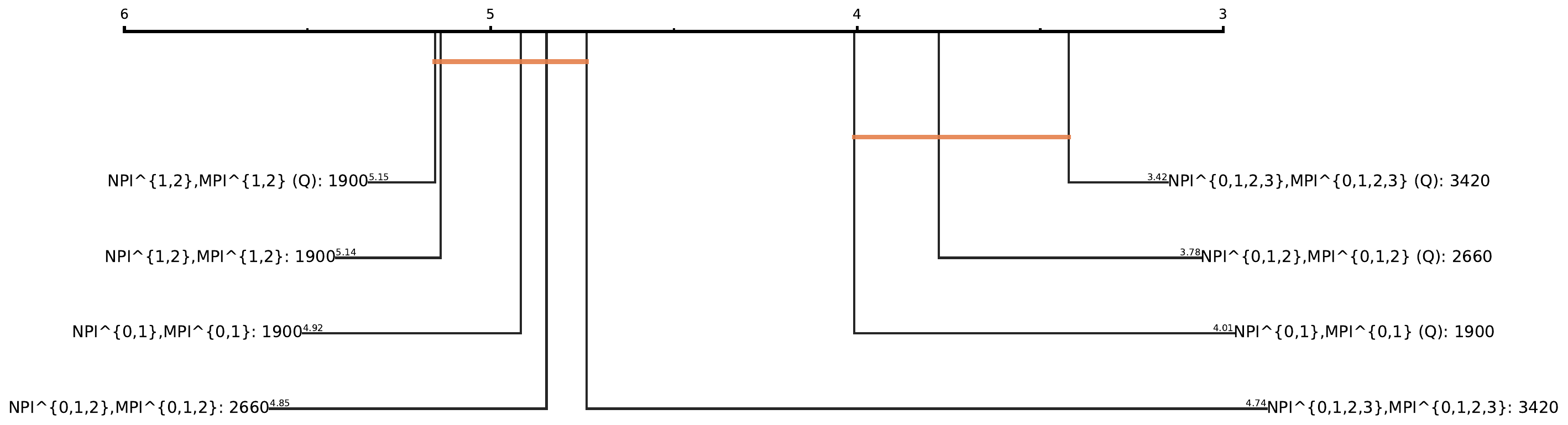}
        \caption{Arctic semiring $\arctic$}
        \label{fig:arctic_sieves_cd}
    \end{minipage}
    \caption*{Critical Difference Diagram of various combinations of feature sieves. Each
        configuration also uses the $\fend$ sieve, which is omitted in the description. Fruits
        marked with a (Q) use quantiles estimated from the training set.}
\end{figure}

Figure~\ref{fig:reals_sieves_cd} compare various combinations of these sieves in $\reals$. All
sieves above combined achieve a large increase in accuracy compared to reduced versions.
Additionally using the third increments $\fnpi^3,\fmpi^3$ doesn't lead to a significantly better
pipeline, while dramatically increasing feature count. A similar result can be observed in
Figure~\ref{fig:arctic_sieves_cd} for $\arctic$. We will here settle for the same feature set.

\subsection{Time warping invariance}\label{sec:exp_time_warping}

The \algoname{Fruits} pipeline built in Section~\ref{sec:general_purpose_fruit} is not time warping
invariant. Several components discussed before are not invariant to stuttering. For $\reals$, we
need to calculate iterated sums on the increments $\finc(x)$ of the input time series. For
$\arctic$, we cannot use $\finc$, as the maximum is already invariant to repetitions of values.

As a weighting control function, we can use $h^{(\text{L1})}$, but not $h^{(\text{id})}$
(Section~\ref{sec:weighting}). A weighting will again only be used for $\reals$. Features of cosine
weighted iterated sums are not time warping invariant.

The only feature sieves discussed so far that are time warping invariant are
$\fnpi^1,\fmpi^1$ and $\fend$. The resulting time-warping invariant pipeline is:
\begin{itemize}[label=$\bullet$]
    \item $\finc\longrightarrow \bigl\{\itsum[\reals,\W]{w}\mid \abs{w}\leq 9\bigr\}
        \longrightarrow\fnpi^1,\fmpi^1,\fend$
    \item $\bigl\{\itsum[\arctic]{w}\mid
        w=\underbrace{[1^{\pm 1}][1^{\mp 1}][1^{\pm 1}]\dotsc}_{\text{length }48}\bigr\}
        \longrightarrow\fnpi^1,\fmpi^1,\fend$
\end{itemize}

Figure~\ref{fig:frequency_sets} shows results of experiments comparing different magnitudes of
stuttering. For that, the training set of any dataset in the UCR archive was left unchanged. The
linear classifier was fit on the computed features and then tested on features of a stuttered
version of the test set. We varied the amount of time points added to the time series. The
positions where stuttering occurred are random. \algoname{Rocket} is not naturally able to deal
with inputs of different lengths in the training and test set. For experiments with
\algoname{Rocket}, we therefore also lengthen the time series in the training set by the same
amount as the test set by repeating the last value a fixed number of times. The results show that
\algoname{Rocket}'s accuracy deteriorates quite quickly even for slightly stuttered time series,
while the features of \algoname{Fruits} are identical in these experiments.

\begin{figure}
    \includegraphics[width=1.0\textwidth]{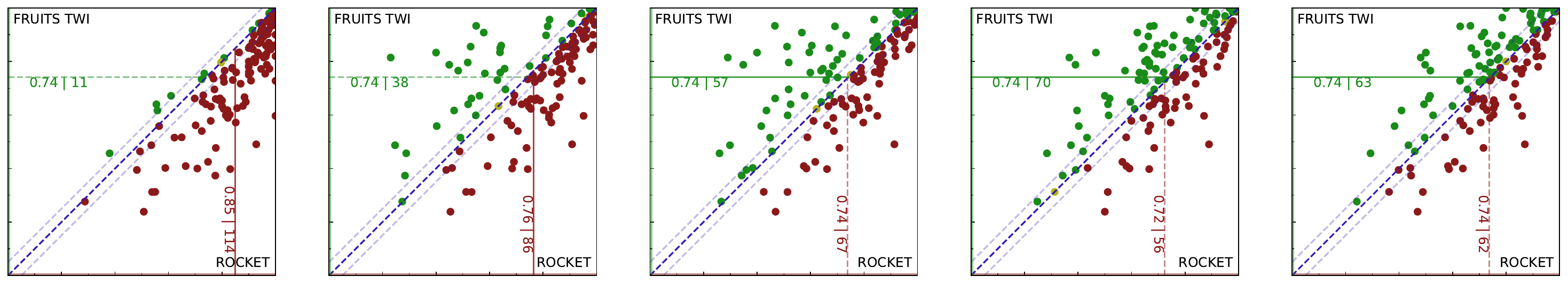}
    \caption{Comparisons of \algoname{Fruits} to \algoname{Rocket} for an increasing magnitude of
        time series stuttering. From left-to-right the proportional additional lengths are
        $0\%,10\%,20\%,50\%$ and $90\%$. FRUITS TWI marks the time warping invariant pipeline.}
    \label{fig:twi_stuttering}
\end{figure}

\paragraph{IMU Data for Point of Impact Localization}

An earlier version of \algoname{Fruits} (only available as an open-source GitHub repository) was
already used to classify a novel dataset in robotics \cite{krieg2022time}. The involved time series
are recordings of an inertial measurement unit (IMU) during collision of a robot with an obstacle.
The recorded variables were linear acceleration and rotational velocity in three dimensions. With
that, one time series is six dimensional. On this dataset, \algoname{Fruits} can compete with
\algoname{Rocket}, even when choosing a small pipeline that is significantly faster than
\algoname{Rocket}. The IMU data is a good example of a practical application of time-warping
invariance. The time of the collision is unknown, and right before it, the robot has nearly
constant acceleration and rotational velocity. This time interval can be interpreted to be a
sequence of stuttered values.

\subsection{A reduced pipeline}

\begin{figure}
    \includegraphics[width=1.0\textwidth]{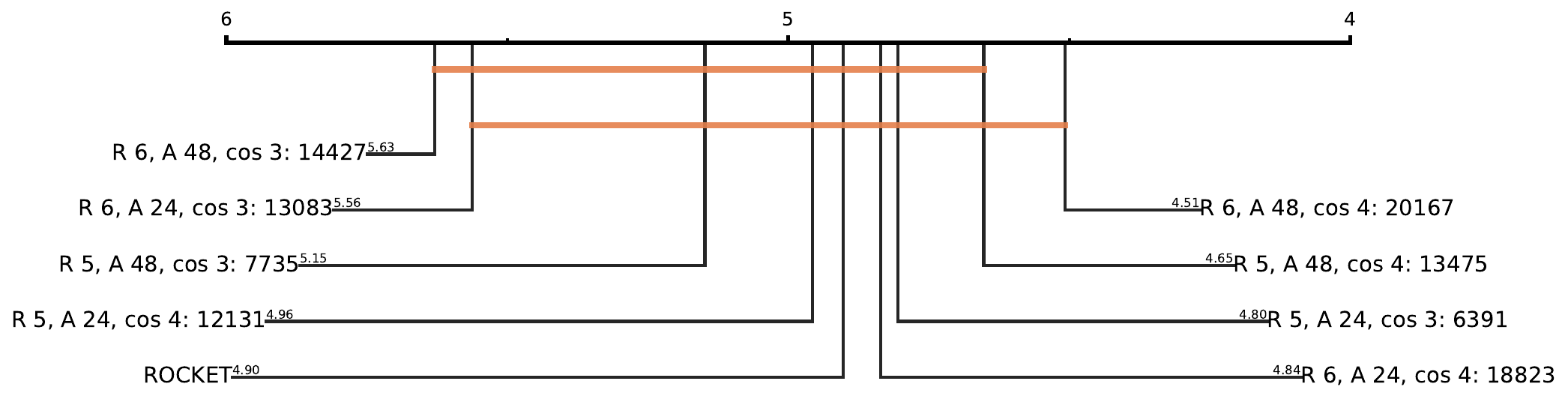}
    \caption{Critical Difference Diagram of fruits with slightly different words for the ISS.
        The name of one configuration tells us the maximum weight of words in $\reals$, the length
        of words in $\arctic$, and the maximum weight used with cosine weighted ISS.
        \algoname{Rocket} is right in the center of this comparison.}
    \label{fig:search_reduced}
\end{figure}

\begin{figure}
    \includegraphics[width=1.0\textwidth]{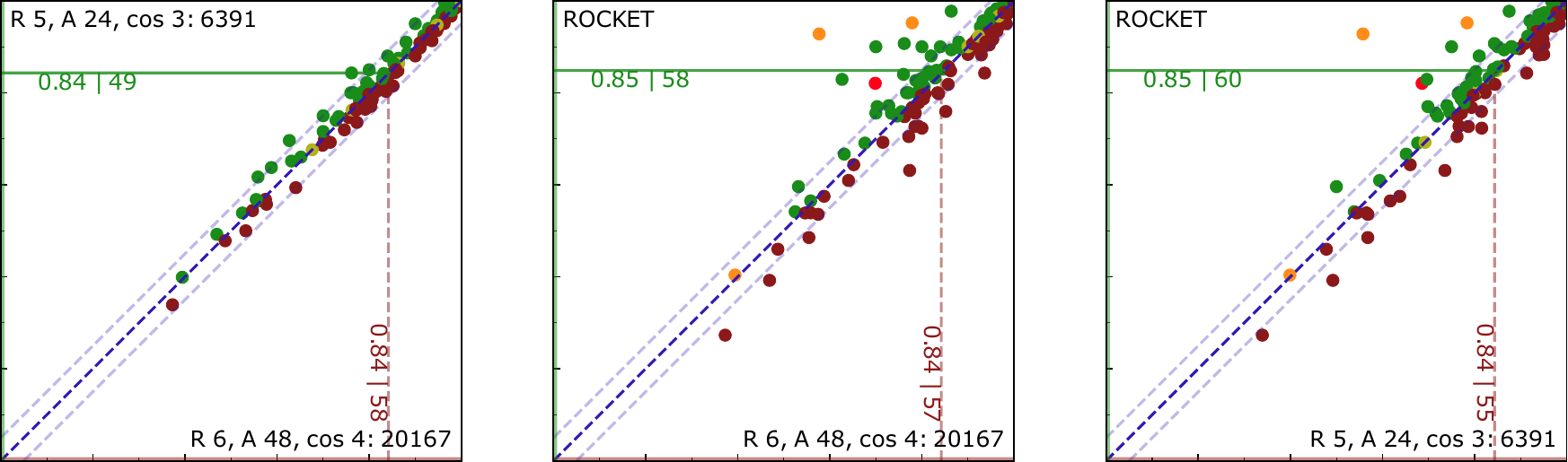}
    \caption{Detailed comparison of fruits from Figure~\ref{fig:search_reduced}. The orange points
    of the two right comparisons mark the datasets PigCVP, PigArtPressure and PigAirwayPressure;
    the bright red point is ChlorineConcentration (see Section~\ref{sec:exp_datasets}).}
    \label{fig:search_reduced_detail}
\end{figure}

Although being very competitive with SOTA methods, the \algoname{Fruits} configuration found in
Section~\ref{sec:general_purpose_fruit} is heavy on number of features and the pipeline is a bit
slower than \algoname{Rocket}. We conducted a wide range of experiments changing the maximum weight
$\abs{w}$ of words $w$ used in this Fruit. Figure~\ref{fig:search_reduced} shows some of the best
alternatives we tested. Interestingly, restricting the weight of words for each type of ISS leads
to a very good pipeline, competing with our choice from Section~\ref{sec:general_purpose_fruit}.
Figure~\ref{fig:search_reduced_detail} presents the scatter plot of \algoname{Rocket}, our general
purpose Fruit and its reduced version.

\section{Outlook}

\begin{itemize}

    \item
    Our python package, \url{https://github.com/irkri/fruits}, is
    object-oriented and modular, thereby easily extensible and readable.
    It uses numba's \cite{lam2015numba} JIT capabilities in parts,
    but is not fully optimized.
    For example, parallelization over training examples is interrupted at each of the three steps
    in \algoname{Fruits}. Grouping them in one function can lead to a significant speed-up.

    \item
    Although the calculation of the argmax indices,
    which we propose here for the first time and whose algorithm
    is given is Algorithm~\ref{alg:argmax},
    leads to a general improvement of classification accuracy,
    we believe that the full potential of this feature has not yet been exploited.

    \item
    The coquantile-features, although theoretically sounds,
    does \emph{not} lead to an improvement of classification accuracy.
    Nonetheless we believe that this feature has potential and
    should be further investigated.
    Similarly, it would be interesting to see if features obtained via the
    Bayesian semiring lead to an improvement of classification accuracy.

    \item
    In this work we have demonstrated that
    the iterated-sums signature can be used as a feature-extraction method
    for time series classification. The natural next step is to use it as a
    \emph{trainable layer} in a neural network.

    \item
    Certain terms of the iterated-sums signature
    are interpretable either as cross-correlations of the time series
    or as certain geometric features of the time series, \cite{diehl2019invariants}.
    It would be interesting to make use of this fact in the context of
    our pipeline.

\end{itemize}

\paragraph*{Acknowledgments}
J. Diehl was supported by a Trilateral ANR-DFG-JST AI program, DFG Project 442590525.
J. Diehl and R. Krieg were supported by the project ``Signatures for Images'' at the Centre for Advanced Study
at the Norwegian Academy of Science and Letters in Oslo, Norway, during the academic year 2023/24.

\printbibliography

\appendix

\section{Iterated-sums signature}\label{app:iss}

\newcommand\e{\mathsf{e}}
Let $A=\{\w1,\dotsc,\w d\}$ be an alphabet whose elements represent dimension indices.
Consider the free commutative monoid $\A$ over it, consisting
of all commutative monomials in the generators $A$. We denote the empty word by $\e$ and, enclose,
for readability, the elements of $\A$ in brackets, e.g. for $d=3$,
\begin{align*}
    \e, [\w 1^3 \w3^5 ], [\w1 \w2 \w3], [\w2^7] \in \A.
\end{align*}

\newcommand\K{\mathds{K}}
For a field $\K$,
we then consider the tensor $\K$-algebra over $\A$
\begin{align*}
    \T(\A)=\bigoplus_{n\geq0}\mathds{K}[\A]^{\otimes n}
          \cong\mathds{K}\e\oplus\left(\bigoplus_{n\geq1}\mathds{K}[\A]^{\otimes n}\right).
\end{align*}
Elements of $\T(\A)$ can be considered to be \emph{finite}, $\mathds K$-linear combinations of words in $\A$.
For example, for $d=3$ and $\K=\R$,
\begin{align*}
    \e \in \R[\A]^{\otimes 0}, \quad
    - [\w 1^3 \w3^5 ] + 17 [\w1 \w2 \w3] + 2 [\w2^7] \in \R[\A]^{\otimes 1}, \quad
    [\w 1][\w 1] - [\w 1^3 \w3^5 ] [\w1 \w2 \w3] \in \R[\A]^{\otimes 2}.
\end{align*}
For a given time series $x\in\tset{d}$ the iterated-sums signature is an element
of the (algebraic) dual space of $\T(\A)$.
We denote the dual pairing with $\scalar{.,.}$.
Then, the signature on the basis elements of words in $\A$ is, for $s \le t$ defined as
\begin{align}
    \label{eq:iss_def}
    \scalar{ [a_1] \dotsm [a_p], \iss{s,t}(x) }
    \coloneqq
    \sum_{\substack{s < t_1< \dotsc < t_p \le t}}
            x_{t_1}^{[a_1]} \dotsm x^{[a_p]}_{t_p}
\end{align}
In the main text we use the short notation
\begin{align*}
    \itsum{w}(x)_t \coloneqq
    \scalar{ [a_1] \dotsm [a_p], \iss{0,t}(x) }.
\end{align*}
\begin{remark}
    Note that the sum ``scans'' over \emph{all} subsequences of $x$ of a fixed length.
    Contrast this with a convolutional neural network (CNN), which only scans over \emph{consecutive} subsequences.
    This point is further discussed in \cite{DEFT2022}.
\end{remark}
For example,
\begin{align*}
    \scalar{ [\w1^2], \iss{0,3}(x) }
    &=
    (x^{[\w1]}_{1})^2 + (x^{[\w1]}_{2})^2 + (x^{[\w1]}_{3})^2 \\
    \scalar{ [\w1^2\w2][\w5], \iss{0,3}(x) }
    &=
    \sum_{0<t_1< t_2 \le 3} x_{t_1}^{[\w1^2 \w2]} x_{t_2}^{[\w5]} \\
    &=
    \sum_{0<t_1 < t_2 \le 3} (x_{t_1}^{[\w1]})^2 x_{t_1}^{[\w2]} x_{t_2}^{[\w5]} \\
    &=
    x_1^{[\w1]} x_1^{[\w2]} x_2^{[\w5]} + ( x_1^{[\w1]} x_1^{[\w2]} + x_2^{[\w1]} x_2^{[\w2]}) x_3^{[\w5]}.
\end{align*}

\begin{theorem}~
    \label{thm:iss}
    \begin{enumerate}
        \item (Invariance)
            The iterated-sums signature is invariant to insertion of zeros:
            let $t \le T$ be any timepoint and let $y$ be the time series
            \begin{align*}
                y_i \coloneqq \left\{\begin{array}{rl}
                    x_i, & i < t\\
                    0, & i=t \\
                    x_{i-1}, & t < i.
                \end{array}\right.
            \end{align*}
            Then $\iss{0,T}(x) = \iss{0,T+1}(y)$.
            In particular, if $x$ has finite support,
            $\iss{0,\infty}(x) = \iss{0,\infty}(y)$.

        \item (Dynamic programming)
        For any word $[a_1]\dotsc [a_p]$,
        \begin{align*}
            \scalar{[a_1]\dotsc[a_p],\iss{s,t}(x)}&
                =\sum_{i=s+1}^t\scalar{[a_1]\cdots[a_{p-1}],\iss{n,i-1}(x)}x_i^{[a_p]}.
        \end{align*}

        \item (Quasi-shuffle identity)
        There is a commutative product $\star$ on $\T(\A)$ such that
        for all elements $\phi, \psi \in \T(\A)$ and all $x \in \tset{d}$,
        and all $s \le t$,
        \begin{align*}
            \scalar{\phi, \iss{s,t}(x)} \scalar{\psi, \iss{s,t}(x)}
            =
            \scalar{\phi \star \psi, \iss{s,t}(x)}.
        \end{align*}

    \end{enumerate}
\end{theorem}

\begin{remark}~
    \begin{enumerate}
        \item
            The ostensibly polynomial-time algorithm for computing a term in the iterated-sums signature
            (after all, \eqref{eq:iss_def} contains $\binom{T}p$ terms),
            is reduced to a linear time algorithm by the dynamic programming property.

        \item
            The quasi-shuffle identity implies that any \emph{polynomial}
            expression in terms of the iterated-sums signature can be re-expressed
            as a \emph{linear} expression in (other) terms of the signature.

    \end{enumerate}

\end{remark}

\section{Argmax indices}
\label{app:argmax_indices}

A pseudo-implementation of the algorithm computing
$\itsumu[\arctic]{w}$, and the corresponding cumulative argmax and argmin,
is shown in Algorithm~\ref{alg:argmax}.
It is best understood by an example.
Let
\begin{align*}
    z = (1,3,-4,2,0,5,1,1),\  w = [1][1^{-1}][1],
\end{align*}
and consider the steps of the first for-loop,
\begin{align*}
    \pmb{k} &\pmb{= 1}                      & \pmb{k} &\pmb{= 2} \\
        z  &=  (1,3,3,3,3,5,5,5)        & z   &= (0,0,7,1,3,0,4,4) \\
    J^{(1)} &= (1,2,2,2,2,6,6,6)        & J^{(2)} &= (1,1,3,3,3,3,3,3) \\
    ~\\
    \pmb{k} &\pmb{= 3} \\
        z   &= (1,3,3,3,3,5,5,5) \\
    J^{(3)} &= (1,2,2,2,2,6,6,6).
\end{align*}
We see that $(J^{(1)}_8, J^{(2)}_8, J^{(3)}_8) = (6,3,6)$, which is \emph{not}
the correct positions of the argmax of $\itsumu[\arctic]{w}(z)$ (they are not even
ordered correctly). The second for-loop corrects this,
\begin{align*}
   \pmb{k} &\pmb{= 3}                                        & \pmb{k} &\pmb{= 2} \\
   J^{(2)} &= (1,1,3,3,3,3,3,3) \text{\ \  (no change)}       & J^{(1)} &= (1,2,2,2,2,2,2,2).
\end{align*}
Now at each timepoint $i$, $(J^{(1)}_i, J^{(2)}_i, J^{(3)}_i)$ contains the correct
tuple giving the argmax (up to $i$) of the expression $z_{t_1} - z_{t_2} + z_{t_3}$.

\begin{algorithm}
    \caption{Implementation of the Arctic Iterated Sum with Argmax Computation}
    \begin{algorithmic}
        \Require $x\in\tset[T]{d}, w=[a_1]\dotsc[a_p]$
        \State $z\gets(\zero,\zero,\dotsc,\zero)\in\tset[T]{d}$
        \State $J^{(1)},\dotsc,J^{(p)}\gets(1,1,\dotsc,1)\in\N^T$
        \For{$k=1,2,\dotsc,p$}
            \State $z\gets z\odot x^{\odot [a_k]}$
            \For{$t=2,3,\dotsc,T$}
                \If{$z_{t-1}\geq z_{t}$}
                    \State $z_t\gets z_{t-1}$
                    \State $J^{(k)}_t\gets J^{(k)}_{t-1}$
                \Else
                    \State $J^{(k)}_t\gets i$
                \EndIf
            \EndFor
        \EndFor
        \For{$k=p,p-1,\dotsc,2$} \Comment{Retranslate argmax of prior steps}
            \State $\hat t\gets J^{(k)}_T$
            \For{$t=\hat t+1,\hat t+2,\dotsc,T$}
                \State $J^{(k-1)}_t\gets J^{(k-1)}_{\hat t}$
            \EndFor
        \EndFor
        \State \Return $J^{(1)}, \dotsc, J^{(p)}$
    \end{algorithmic}
    \label{alg:argmax}
\end{algorithm}

\section{Detailed results on the UCR archive}

\begin{scriptsize}
    \csvreader[
        longtable=l|r|r|r|r|r|r,
        table head=\caption{%
            Results of our general pipeline from Section~\ref{sec:general_purpose_fruit}, its
            reduced Version and \algoname{Rocket}. All values are rounded to two decimal places.
            We ran the experiments for \algoname{Rocket} on comparable hardware using the
            implementation in the Python package ``sktime'' \cite{loning2019sktime}.
            }\label{tbl:ucr_results}\\\hline
        Datensatz & \makecell{\algoname{Fruits} (general)\\Time in s} &
        \makecell{\algoname{Fruits} (general)\\Accuracy} & \makecell{\algoname{Fruits} (reduced)
        \\Time in s} & \makecell{\algoname{Fruits} (reduced)\\Accuracy} &
        \makecell{\algoname{Rocket}\\Time in s} & \makecell{\algoname{Rocket}\\Accuracy}\\\hline%
        \endfirsthead%
        \endhead\bottomrule%
        \multicolumn{5}{c}{$\dotsc$}\endfoot\endlastfoot,
        late after line=\\ ,
        late after first line=\\ ,
        late after last line=\\\hline,
    ]%
    {tables/UCR.csv}%
    {2=\Dataset,3=\GeneralTime,4=\ReducedTime,5=\ROCKETTime,6=\General,7=\Reduced,8=\ROCKET}%
    {\Dataset & \GeneralTime & \General & \ReducedTime & \Reduced & \ROCKETTime & \ROCKET}
\end{scriptsize}

\end{document}